\documentclass{article}

    \PassOptionsToPackage{numbers, compress}{natbib}
\usepackage[final]{neurips_2019}
\usepackage[utf8]{inputenc} % allow utf-8 input
\usepackage[T1]{fontenc}    % use 8-bit T1 fonts
\usepackage{booktabs}       % professional-quality tables
\usepackage{amsfonts}       % blackboard math symbols
\usepackage{nicefrac}       % compact symbols for 1/2, etc.
\usepackage{microtype}      % microtypography
\PassOptionsToPackage{numbers, compress}{natbib}

\usepackage{wrapfig}

\usepackage[table]{xcolor}
\definecolor{mydarkred}{rgb}{0.6,0,0}
\definecolor{mydarkgreen}{rgb}{0,0.6,0}
\usepackage[colorlinks,
linkcolor=mydarkred,
citecolor=mydarkgreen]{hyperref}
\usepackage{url}

\usepackage{booktabs}       % professional-quality tables
\usepackage{amsfonts} % blackboard math symbols
\usepackage{nicefrac}       % compact symbols for 1/2, etc.
\usepackage{microtype}      % microtypography
\usepackage{algorithmic}
\usepackage{algorithm}
\usepackage[algo2e,linesnumbered]{algorithm2e}
\usepackage{graphicx}
\usepackage{multirow}
\usepackage{epstopdf}
\usepackage{multicol}

\usepackage{caption}
\usepackage{subfigure}
\usepackage{enumitem}
\usepackage{amsmath}
\usepackage{pifont}
\newtheorem{theorem}{Theorem}
\newtheorem{lemma}{Lemma}
\usepackage{amssymb}
\usepackage{pifont}

\title{Are Anchor Points Really Indispensable\\ in Label-Noise Learning?}

\author{
  Xiaobo Xia$^{1,2}$\quad
  Tongliang Liu$^{1}$\quad
  Nannan Wang$^2$\\
  \textbf{Bo Han$^3$\quad Chen Gong$^4$\quad Gang Niu$^3$\quad Masashi Sugiyama$^{3,5}$}\\[1ex]
  $^1$University of Sydney\quad
  $^2$Xidian University\quad
  $^3$RIKEN\\
  $^4$Nanjing University of Science and Technology\quad
  $^5$University of Tokyo\\
}

\begin{document}

\maketitle

\begin{abstract}
In label-noise learning, the \textit{noise transition matrix}, denoting the probabilities that clean labels flip into noisy labels, plays a central role in building \textit{statistically consistent classifiers}. Existing theories have shown that the transition matrix can be learned by exploiting \textit{anchor points} (i.e., data points that belong to a specific class almost surely). However, when there are no anchor points, the transition matrix will be poorly learned, and those previously consistent classifiers will significantly degenerate.
In this paper, without employing anchor points, we propose a \textit{transition-revision} ($T$-Revision) method to effectively learn transition matrices, leading to better classifiers. Specifically, to learn a transition matrix, we first initialize it by exploiting data points that are similar to anchor points, having high \textit{noisy class posterior probabilities}. Then, we modify the initialized matrix by adding a \textit{slack variable}, which can be learned and validated together with the classifier by using noisy data. Empirical results on benchmark-simulated and real-world label-noise datasets demonstrate that without using exact anchor points, the proposed method is superior to state-of-the-art label-noise learning methods.
\end{abstract}
\section{Introduction}
Label-noise learning can be dated back to \cite{angluin1988learning} but becomes a more and more important topic recently. The reason is that, in this era, datasets are becoming bigger and bigger. Often, large-scale datasets are infeasible to be annotated accurately due to the expensive cost, which naturally brings us cheap datasets with noisy labels.

Existing methods for label-noise learning can be generally divided into two categories: algorithms that result in \textit{statistically inconsistent/consistent} classifiers. Methods in the first category usually employ heuristics to reduce the side-effect of noisy labels. For example, many state-of-the-art approaches in this category are specifically designed to, e.g., select reliable examples \cite{yu2019does,han2018co,malach2017decoupling}, reweight examples \cite{ren2018learning,jiang2018mentornet}, correct labels \cite{ma2018dimensionality,kremer2018robust,tanaka2018joint,reed2014training}, employ side information \cite{vahdat2017toward,li2017learning}, and (implicitly) add regularization \cite{han2018masking,guo2018curriculumnet,veit2017learning,vahdat2017toward,li2017learning}.
All those methods were reported to work empirically very well. However, the differences between the learned classifiers and the optimal ones for clean data are not guaranteed to vanish, i.e., no statistical consistency has been guaranteed.

The above issue motivates researchers to explore algorithms in the second category: \textit{risk-/classifier-consistent} algorithms. In general, risk-consistent methods possess statistically consistent estimators to the clean risk (i.e., risk w.r.t.~the clean data), while classifier-consistent methods guarantee the classifier learned from the noisy data is consistent to the optimal classifier (i.e., the minimizer of the clean risk) \cite{vapnik2013nature}.
Methods in this category utilize the \textit{noise transition matrix}, denoting the probabilities that clean labels flip into noisy labels, to build consistent algorithms. 
Let ${Y}$ denote the variable for the clean label, $\bar{Y}$ the noisy label, and $X$ the instance/feature. The basic idea is that given the \textit{noisy class posterior probability} $P({\bf \bar{Y}}|X=x)=[P(\bar{Y}=1|X=x),\ldots,P(\bar{Y}=C|X=x)]^\top$ (which can be learned using noisy data) and the transition matrix $T(X=x)$ where $T_{ij}(X=x)=P(\bar{Y}=j|Y=i,X=x)$, the \textit{clean class posterior probability} $P({\bf Y}|X=x)$ can be inferred, i.e., $P({\bf Y}|X=x)=(T(X=x)^\top)^{-1}P({\bf \bar{Y}}|X=x)$. 
For example, loss functions are modified to ensure risk consistency, e.g., \cite{zhang2018generalized,kremer2018robust,liu2016classification,northcuttlearning,scott2015rate,natarajan2013learning}; a noise adaptation layer is added to deep neural networks to design classifier-consistent deep learning algorithms \cite{goldberger2016training,patrini2017making,thekumparampil2018robustness,yu2018learning}. Those algorithms are strongly theoretically grounded but heavily rely on the success of learning transition matrices.

Given risk-consistent estimators, one stream to learn the transition matrix is the \textit{cross-validation} method (using only noisy data) for binary classification \cite{natarajan2013learning}. However, it is prohibited for multi-class problems as its computational complexity grows exponentially to the number of classes. 
Besides, the current risk-consistent estimators involve the inverse of the transition matrix, making tuning the transition matrix inefficient and also leading to performance degeneration \cite{patrini2017making}, especially when the transition matrix is non-invertible. 
Independent of risk-consistent estimators, another stream to learn the transition matrix is closely related to \textit{mixture proportion estimation} \cite{vandermeulen2016operator}. A series of assumptions \cite{scott2013classification,liu2016classification,scott2015rate,ramaswamy2016mixture} were proposed to efficiently learn transition matrices (or mixture parameters) by only exploiting the noisy data.
All those assumptions require anchor points, i.e., instances belonging to a specific class with probability exactly one or close to one. Nonetheless, without anchor points, the transition matrix could be poorly learned, which will degenerate the accuracies of existing consistent algorithms.

Therefore, in this paper, to handle the applications where the anchor-point assumptions are violated \cite{yu2018efficient,vandermeulen2019operator}, we propose a \textit{transition-revision} ($T$-Revision) method to effectively learn transition matrices, leading to better classifiers. In a high level, we design a deep-learning-based risk-consistent estimator to tune the transition matrix accurately. Specifically, we first initialize the transition matrix by exploiting examples that are similar to anchor points, namely, those having high estimated \textit{noisy class posterior probabilities}. Then, we modify the initial matrix by adding a \textit{slack variable}, which will be learned and validated together with the classifier by using noisy data only. Note that given true transition matrix, the proposed estimator will converge to the classification risk w.r.t. clean data by increasing the size of noisy training examples. Our heuristic for tuning the transition matrix is that a favorable transition matrix would make the classification risk w.r.t. clean data small. We empirically show that the proposed $T$-Revision method will enable tuned transition matrices to be closer to the ground truths, which explains why $T$-Revision is much superior to state-of-the-art algorithms in classification. 

The rest of the paper is organized as follows. In Section \ref{sec:with-anchor} we review label-noise learning with anchor points. In Section \ref{sec:without-anchor}, we discuss how to learn the transition matrix and classifier without anchor points. Experimental results are provided in Section \ref{sec:experiment}. Finally, we conclude the paper in Section \ref{sec:conclusion}.

\section{Label-Noise Learning with Anchor Points }\label{sec:with-anchor}

In this section, we briefly review label-noise learning when there are anchor points.

\textbf{Preliminaries}\ \
Let $D$ be the distribution of a pair of random variables $(X,Y)\in \mathcal{X}\times\{1,2,\ldots,C\}$, where the feature space $\mathcal{X}\subseteq\mathbb{R}^d$ and $C$ is the size of label classes. Our goal is to predict a label $y$ for any given instance $x\in\mathcal{X}$. However, in many real-world classification problems, training examples drawn independently from distribution $D$ are unavailable. Before being observed, their true labels are independently flipped and what we can obtain is a noisy training sample $\{(X_i,\bar{Y}_i)\}_{i=1}^{n}$, where $\bar{Y}$ denotes the noisy label. Let $\bar{D}$ be the distribution of the noisy random variables $(X,\bar{Y})\in \mathcal{X}\times\{1,2,\ldots,C\}$.

\textbf{Transition matrix}\ \
The random variables $\bar{Y}$ and $Y$ are related through a \textit{noise transition matrix} $T\in[0,1]^{C\times C}$ \cite{cheng2017learning}. Generally, the transition matrix depends on instances, i.e., $T_{ij}(X=x)=P(\bar{Y}=j|Y=i,X=x)$. Given only noisy examples, the \textit{instance-dependent} transition matrix is \textit{non-identifiable} without any additional assumption. For example, 
$P(\bar{Y}=j|X=x)=\sum_{i=1}^C T_{ij}(X=x)P(Y=i|X=x)=\sum_{i=1}^C T'_{ij}(X=x)P'(Y=i|X=x)$ are both valid,
when $T'_{ij}(X=x)=T_{ij}(X=x)P({Y}=i|X=x)/P'(\bar{Y}=i|X=x)$. In this paper, we study the  \textit{class-dependent} and \textit{instance-independent} transition matrix,  i.e., $P(\bar{Y}=j|Y=i,X=x)= P(\bar{Y}=j|Y=i)$, which is identifiable under mild conditions and on which the vast majority of current methods focus \cite{han2018co,han2018masking,patrini2017making,northcuttlearning,natarajan2013learning}.

\textbf{Consistent algorithms}\ \
The transition matrix bridges the class posterior probabilities for noisy and clean data, i.e., $P(\bar{Y}=j|X=x)=\sum_{i=1}^C T_{ij}P(Y=i|X=x)$. Thus, it has been exploited to build consistent algorithms. Specifically, it has been used to modify loss functions to build \textit{risk-consistent} estimators, e.g., \cite{natarajan2013learning,scott2015rate,patrini2017making}, and has been used to correct hypotheses to build \textit{classifier-consistent} algorithms, e.g., \cite{goldberger2016training,patrini2017making,yu2018learning}. Note that an estimator is risk-consistent if, by increasing the size of noisy examples, the \textit{empirical risk} calculated by noisy examples and the modified loss function will converge to the \textit{expected risk} calculated by clean examples and the original loss function. Similarly, an algorithm is classifier-consistent if, by increasing the size of noisy examples, the learned classifier will converge to the optimal classifier learned by clean examples. Definitions of the expected and empirical risks can be found in Appendix B, where we further discuss how consistent algorithms work.

\textbf{Anchor points}\ \
The successes of consistent algorithms rely on firm bridges, i.e., accurately learned transition matrices. To learn transition matrices, the concept of anchor point was proposed \cite{liu2016classification,scott2015rate}. Anchor points are defined in the clean data domain, i.e., an instance $x$ is an anchor point for the class $i$ if $P(Y=i|X=x)$ is equal to one or close to one\footnote{In the literature, the assumption $\inf_{x}P(Y=i|X=x)\to 1$ was introduced as irreducibility \cite{blanchard2010semi} to ensure the transition matrix is identifiable; an anchor point $x$ for class $i$ is defined by $P(Y=i|X=x)=1$ \cite{scott2015rate,liu2016classification} to ensure a fast convergence rate. In this paper, we generalize the definition for the anchor point family, including instances whose class posterior probability $P(Y=i|X=x)$ is equal to or close to one.}. Given an $x$, if $P(Y=i|X=x)=1$, we have that for $k\neq i, P(Y=k|X=x)=0$. Then, we have
\begin{align}
\label{eq:anchor-estim}%
P(\bar{Y}=j|X=x)=\sum_{k=1}^C T_{kj}P(Y=k|X=x)=T_{ij}.
\end{align}
Namely, $T$ can be obtained via estimating the noisy class posterior probabilities for anchor points \cite{yu2018learning}. However, the requirement of given anchor points is a bit strong. Thus, anchor points are assumed to exist but unknown in datasets, which can be identified either theoretically \cite{liu2016classification} or heuristically \cite{patrini2017making}.

\begin{wrapfigure}{r}{19em}
\centering
\includegraphics[width=0.2\textwidth]{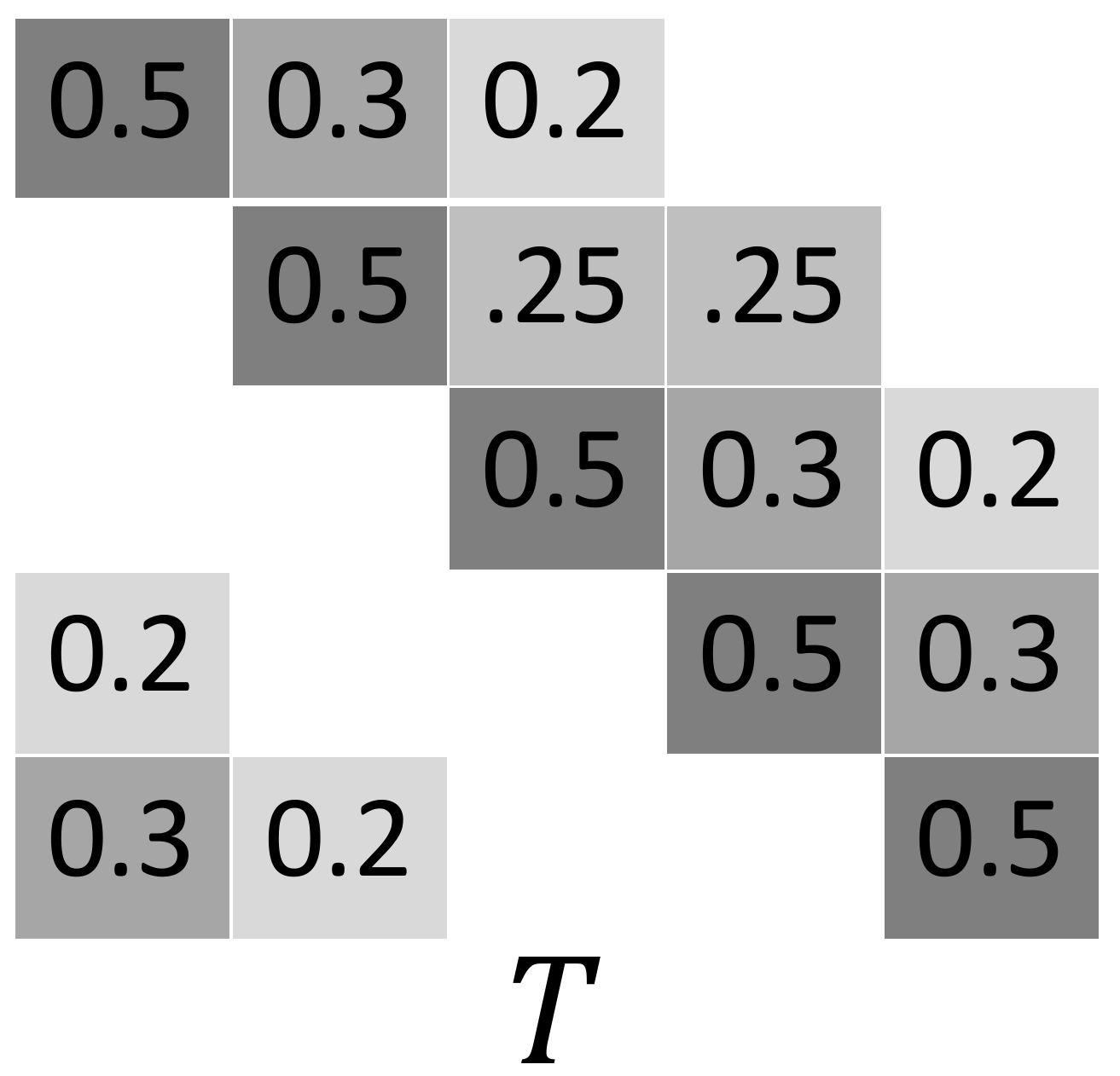}
\qquad
\includegraphics[width=0.2\textwidth]{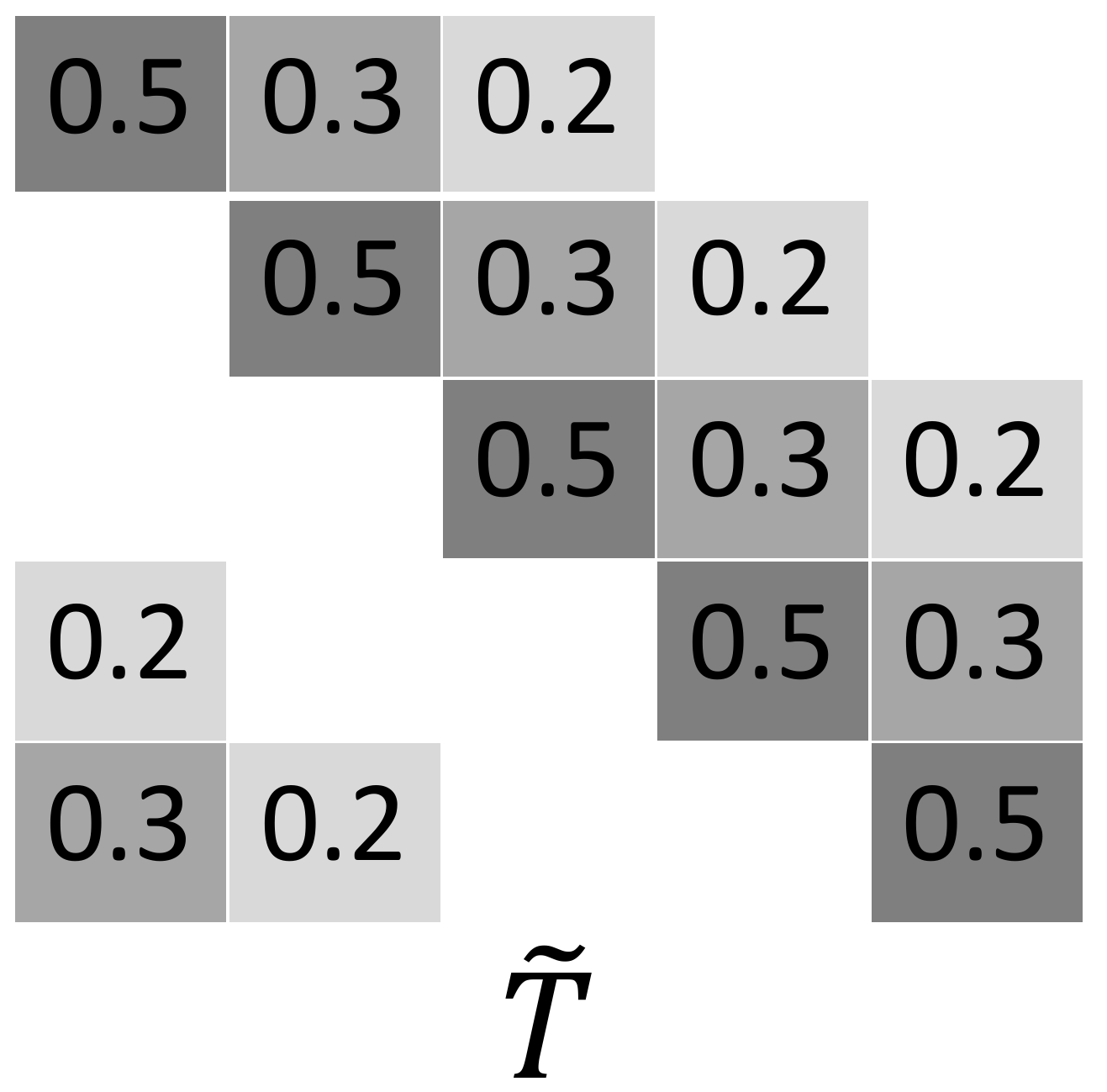}
\caption{Illustrative experimental results (using a 5-class classification problem as an example). The noisy class posterior probability $P({\bf \bar{Y}}|X=x)$ can be estimated by exploiting noisy data. Let an example have $P({\bf \bar{Y}}|X=x)=[0.141; 0.189; 0.239;0.281;0.15]$. If the true transition matrix $T$ is given, we can infer the clean class posterior probability as $P({\bf {Y}}|X=x)=(T^\top)^{-1}P({\bf \bar{Y}}|X=x)=[0.15;0.28;0.25;0.3;0.02]$ and that the instance belongs to the fourth class. However, if the transition matrix is not accurately learned as $\tilde{T}$ (only slightly differing from $T$ with two entries in the second row), the clean class posterior probability can be inferred as $P({\bf{Y}}|X=x)=(\tilde{T}^\top)^{-1}P({\bf \bar{Y}}|X=x)=[0.1587; 0.2697; 0.2796; 0.2593; 0.0325]$ and the instance could be mistakenly classified  into the third class.}
\label{fig:matrix}
\vspace{-60px}
\end{wrapfigure}

Transition matrix learning is also closely related to \textit{mixture proportion estimation} \cite{vandermeulen2016operator}, which is independent of classification. By giving only noisy data, to ensure the \textit{learnability} and efficiency of learning transition matrices (or mixture parameters), a series of assumptions were proposed, 
e.g., \textit{irreducibility} \cite{scott2013classification}, \textit{anchor point} \cite{liu2016classification,scott2015rate}, and \textit{separability} \cite{ramaswamy2016mixture}. All those assumptions require anchor points or instances belonging to a specific class with probability one or approaching one.

When there are no anchor points in datasets/data distributions, all the above mentioned methods will lead to inaccurate transition matrices, which will degenerate the performances of current consistent algorithms. This motivates us to investigate how to maintain the efficacy of those consistent algorithms without using exact anchor points.

\section{Label-Noise Learning without Anchor Points}\label{sec:without-anchor}

This section presents a deep-learning-based risk-consistent estimator for the classification risk w.r.t. clean data. We employ this estimator to tune the transition matrix effectively without using anchor points, which finally leads to better classifiers.

\subsection{Motivation}
According to Eq.~\eqref{eq:anchor-estim}, to learn the transition matrix, $P({\bf \bar{Y}}|X=x)$ needs to be estimated and anchor points need to be given. Note that learning $P({\bf \bar{Y}}|X=x)$ may introduce error. Even worse, when there are no anchor points, it will be problematic if we use existing methods \cite{scott2013classification,liu2016classification,scott2015rate,ramaswamy2016mixture} to learn transition matrices. For example, let $P({\bf {Y}}|X=x^i)$ be the $i$-th column of a matrix $L$, $i=1,\ldots,C$. If $x^i$ is an anchor point for the $i$-th class, then $L$ is an identity matrix. According to Eq.~(\ref{eq:anchor-estim}), if we use $x^i$ as an anchor point for the $i$-th class while $P(Y=i|X=x^i)\neq 1$ (e.g., the identified instances in \cite{patrini2017making} are not guaranteed to be anchor points), the learned transition matrix would be $TL$, where $L$ is a non-identity matrix. This means that transition matrices will be inaccurately 
estimated.

Based on inaccurate transition matrices, the accuracy of current consistent algorithms will significantly degenerate.
To demonstrate this, Figure \ref{fig:matrix} shows that given a noisy class posterior probability $P({\bf \bar{Y}}|X=x)$, even if the transition matrix changes slightly by two entries, e.g., $\|T-\tilde{T}\|_1/\|T\|_1=0.02$ where $T$ and $\tilde{T}$ are defined in Figure \ref{fig:matrix} and $\|T\|_1=\sum_{ij}|T_{ij}|$, the inferred class posterior probability for the clean data may lead to an incorrect classification. Since anchor points require clean class posterior probabilities to be or approach one, which is quite strong to some real-world applications \cite{yu2018efficient,vandermeulen2019operator}, we would like to study how to maintain the performances of current consistent algorithms when there are no anchor points and then transition matrices are inaccurately learned.

\subsection{Risk-consistent estimator}
Intuitively, the entries of transition matrix can be tuned by minimizing the risk-consistent estimator, since the estimator is asymptotically identical to the expected risk for the clean data and that a favorable transition matrix should make the clean expected risk small. However, existing risk-consistent estimators involve the inverse of transition matrix (more details are provided in Appendix B), which degenerates classification performances \cite{patrini2017making} and makes tuning the transition matrix ineffectively.
To address this, we propose a risk-consistent estimator that does not involve the inverse of the transition matrix.

\begin{figure}[!tp]
\centering
\includegraphics[width=0.96\textwidth]{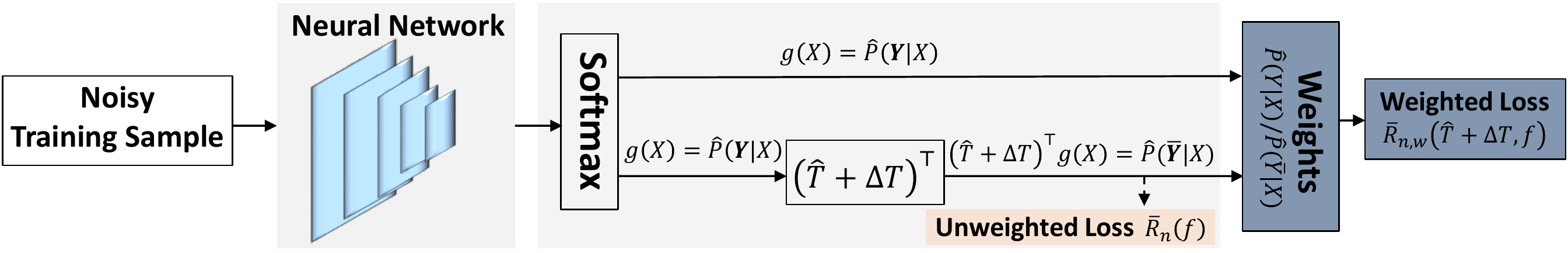}
\caption{An overview of the proposed method. The proposed method will learn a more accurate classifier because the transition matrix is renovated.}
\label{fig:flowchart}
\end{figure}

The inverse of transition matrix is involved in risk-consistent estimators, since the noisy class posterior probability $P({\bf \bar{Y}}|X=x)$ and the transition matrix are explicitly or implicitly used to infer the clean class posterior probability $P({\bf {Y}}|X=x)$, i.e., $P({\bf {Y}}|X=x)=(T^\top)^{-1}P({\bf \bar{Y}}|X=x)$. To avoid the inverse in building risk-consistent estimators, we directly estimate $P({\bf {Y}}|X=x)$ instead of inferring it through $P({\bf \bar{Y}}|X=x)$. Thanks to the equation $T^\top P({\bf {Y}}|X=x)=P({\bf \bar{Y}}|X=x)$, $P({\bf {Y}}|X=x)$ and $P({\bf \bar{Y}}|X=x)$ could be estimated at the same time by adding the true transition matrix to modify the output of the softmax function, e.g., \cite{yu2018learning,patrini2017making}. Specifically, ${P}({\bf \bar{Y}}|X=x)$ can be learned by exploiting the noisy data, as shown in Figure \ref{fig:flowchart} by minimizing the unweighted loss $\bar{R}_{n}(f)=1/n\sum_{i=1}^{n}\ell(f(X_i),\bar{Y}_i)$, where $\ell(f(X),\bar{Y})$ is a \textit{loss function} \cite{mohri2018foundations}.
Let $\hat{T}+\Delta T$ be the true transition matrix, i.e., $\hat{T}+\Delta T=T$. {Due to $P({\bf \bar{Y}}|X=x)=T^\top P({\bf {Y}}|X=x)$, the output of the softmax function $g(x)=\hat{P}({\bf {Y}}|X=x)$ before the transition matrix is an approximation for $P({\bf {Y}}|X=x)$. However, the learned $g(x)=\hat{P}({\bf {Y}}|X=x)$ by minimizing the unweighted loss may perform poorly if the true transition matrix is inaccurately learned as explained in the motivation.}

If having $P({\bf {Y}}|X=x)$ and $P({\bf \bar{Y}}|X=x)$, we could employ the \textit{importance reweighting} technique \cite{gretton2009covariate,liu2016classification} to rewrite the expected risk w.r.t. clean data without involving the inverse of transition matrix.
Specifically, 
\begin{eqnarray}
\label{eq:importance}%
&&R(f)=\mathbb{E}_{(X,Y)\sim D}[\ell(f(X),Y)]=\int_{x}\sum_iP_D(X=x,Y=i)\ell(f(x),i)dx\nonumber\\
&&=\int_{x}\sum_iP_{\bar{D}}(X=x,\bar{Y}=i)\frac{P_D(X=x,\bar{Y}=i)}{P_{\bar{D}}(X=x,\bar{Y}=i)}\ell(f(x),i)dx\nonumber\\
&&=\int_{x}\sum_iP_{\bar{D}}(X=x,\bar{Y}=i)\frac{P_D(\bar{Y}=i|X=x)}{P_{\bar{D}}(\bar{Y}=i|X=x)}\ell(f(x),i)dx\\
&&=\mathbb{E}_{(X,Y)\sim \bar{D}}[\bar{\ell}(f(X),Y)],\nonumber
\end{eqnarray}
where $D$ denotes the distribution for clean data, $\bar{D}$ for noisy data, $\bar{\ell}(f(x),i)=\frac{P_D(\bar{Y}=i|X=x)}{P_{\bar{D}}(\bar{Y}=i|X=x)}\ell(f(x),i)$, and the second last equation holds because label noise is assumed to be independent of instances. In the rest of the paper, we have omitted the subscript for $P$ when no confusion is caused. Since $P({\bf \bar{Y}}|X=x)=T^\top P({\bf {Y}}|X=x)$ and that the diagonal entries of (learned) transition matrices for label-noise learning are all much larger than zero, $P_D(\bar{Y}=i|X=x)\neq 0$ implies $P_{\bar{D}}(\bar{Y}=i|X=x)\neq 0$, which also makes the proposed importance reweighting method stable without truncating the importance ratios.

Eq.~(\ref{eq:importance}) shows that the expected risk w.r.t. clean data and the loss $\ell(f(x),i)$ is equivalent to an expected risk w.r.t. noisy data and a reweighted loss, i.e., $\frac{P_D(\bar{Y}=i|X=x)}{P_{\bar{D}}(\bar{Y}=i|X=x)}\ell(f(x),i)$. The empirical counterpart of the risk in the rightmost-hand side of Eq.~(\ref{eq:importance}) is therefore a risk-consistent estimator for label-noise learning.
We exploit a deep neural network to build this counterpart.
As shown in Figure \ref{fig:flowchart}, we use the output of the softmax function $g(x)$ to approximate ${P}({\bf {Y}}|X=x)$, i.e., $g(x)=\hat{P}({\bf {Y}}|X=x)\approx {P}({\bf {Y}}|X=x)$. Then, $T^\top g(x)$ (or $(\hat{T}+\Delta T)^\top g(x)$ in the figure) is an approximation for ${P}({\bf \bar{Y}}|X=x)$, i.e., $T^\top g(x)=\hat{P}({\bf \bar{Y}}|X=x)\approx {P}({\bf \bar{Y}}|X=x)$.
By employing $\hat{P}(Y=y|X=x)/\hat{P}(\bar{Y}=y|X=x)$ as weight, we build the risk-consistent estimator as 
\begin{align}
\label{eq:em_importance}%
{
\bar{R}_{n,w}(T,f)=\frac{1}{n}\sum_{i=1}^{n}\frac{{g}_{\bar{Y}_i}(X_i)}{(T^\top{g})_{\bar{Y}_i}(X_i)}\ell(f(X_i),\bar{Y}_i),}
\end{align}
where $f(X)=\arg\max_{j\in\{1,\ldots,C\}}{g}_j(X)$, $g_j(X)$ is an estimate for $P(Y=j|X)$, and the subscript $w$ denotes that the loss function is weighted. Note that if the true transition matrix $T$ is given, $\bar{R}_{n,w}(T,f)$ only has one argument $g$ to learn.

\begin{algorithm}[!tp]
 {\bfseries Input}: Noisy training sample $\mathcal{D}_t$; Noisy validation set $\mathcal{D}_v$. 

\textbf{Stage 1: Learn $\hat{T}$}
	
	1: Minimize the unweighted loss to learn $\hat{P}({\bf \bar{Y}}|X=x)$ without a noise adaption layer;

	2: Initialize $\hat{T}$ according to Eq.~(\ref{eq:anchor-estim}) by using instances with the highest $\hat{P}(\bar{Y}=i|X=x)$ as anchor points for the $i$-th class;

\textbf{Stage 2: Learn the classifier $f$ and $\Delta T$}

    3: Initialize the neural network by minimizing the weighted loss with a noisy adaption layer $\hat{T}^\top$;

    4: Minimize the weighted loss to learn $f$ and $\Delta T$ with a noisy adaption layer $(\hat{T}+\Delta T)^\top$; 

	//Stopping criterion for learning $\hat{P}({\bf \bar{Y}}|X=x)$, $f$ and $\Delta T$: when $\hat{P}({\bf \bar{Y}}|X=x)$ yields the minimum classification error on the noisy validation set $\mathcal{D}_v$

{\bfseries Output}: $\hat{T}$, $\Delta T$, and $f$.
\caption{Reweight $T$-Revision (Reweight-R) Algorithm.}
\label{alg:reweighting}
\end{algorithm}

\subsection{Implementation and the $T$-revision method}
When the true transition matrix $T$ is unavailable, we propose to use $\bar{R}_{n,w}(\hat{T}+\Delta T, f)$ to approximate $R(f)$, as shown in Figure \ref{fig:flowchart}. 
To minimize $\bar{R}_{n,w}(\hat{T}+\Delta T,f)$, a two-stage training procedure is proposed. Stage 1:
first learn $P({\bf \bar{Y}}|X=x)$ by minimizing the unweighted loss without a noise adaption layer and initialize $\hat{T}$ by exploiting examples that have the highest learned $\hat{P}({\bf \bar{Y}}|X=x)$; Stage 2: modify the initialization $\hat{T}$ by adding a slack variable $\Delta T$ and learn the classifier and $\Delta T$ by minimizing the weighted loss. The procedure is called the Weighted $T$-Revision method and is summarized in Algorithm~\ref{alg:reweighting}. It is worthwhile to mention that all anchor points based consistent estimators for label-noise learning have a similar two-stage training procedure. Specifically, with one stage to learn ${P}({\bf \bar{Y}}|X=x)$ and the transition matrix and a second stage to learn the classifier for the clean data.

The proposed $T$-revision method works because we learn $\Delta T$ by minimizing the risk-consistent estimator, which is asymptotically equal to the expected risk w.r.t. clean data. The learned slack variable can also be validated on the noisy validation set, i.e., to check if $\hat{P}({\bf \bar{Y}}|X=x)$ fits the validation set. The philosophy of our approach is similar to that of the cross-validation method. However, the proposed method does not need to try different combinations of parameters ($\Delta T$ is learned) and thus is much more computationally efficient. Note that the proposed method will also boost the performances of consistent algorithms even there are anchor points as the transition matrices and classifiers are jointly learned. Note also that if a clean validation set is available, it can be used to better initialize the transition matrix, to better validate the slack variable $\Delta T$, and to fine-tune the deep network.

\subsection{Generalization error}
While we have discussed the use of the proposed estimator for evaluating the risk w.r.t clean data, we theoretically justify how it generalizes for learning classifiers. 
Assume the neural network has $d$ layers, parameter matrices $W_1,\ldots,W_d$, and activation functions $\sigma_1,\ldots,\sigma_{d-1}$for each layer. Let denote the mapping of the neural network by $h: x\mapsto W_d\sigma_{d-1}(W_{d-1}\sigma_{d-2}(\ldots \sigma_1(W_1x)))\in\mathbb{R}^C$. Then, the output of the softmax is defined by $g_i(x)=\exp{(h_i(x))}/\sum_{k=1}^{C}\exp{(h_k(x))}, i=1,\ldots,C$.
Let $\hat{f}=\arg\max_{i\in\{1,\ldots,C\}}\hat{g}_i$ be the classifier learned from the hypothesis space $F$ determined by the real-valued parameters of the neural network, i.e., $\hat{f}=\arg\min_{f\in F}\bar{R}_{n,w}(f)$.

To derive a generalization bound, as the common practice \cite{boucheron2005theory,mohri2018foundations}, we assume that instances are upper bounded by $B$, i.e., $\|x\|\leq B$ for all $x\in\mathcal{X}$, and that the loss function is $L$-Lipschitz continuous w.r.t. $f(x)$ and upper bounded by $M$, i.e., for any $f_1,f_2 \in F$ and any $(x,\bar{y})$, $|\ell(f_1(x),\bar{y})-\ell(f_2(x),\bar{y})|\leq L|f_1(x)-f_2(x)|$, and for any $(x,\bar{y})$, $\ell(f(x),\bar{y})\leq M$. 
\begin{theorem}\label{thm:main}
Assume the Frobenius norm of the weight matrices $W_1,\ldots,W_d$ are at most $M_1,\ldots, M_d$. Let the activation functions be 1-Lipschitz, positive-homogeneous, and applied element-wise (such as the ReLU).
Let the loss function be the \textit{cross-entropy loss}, i.e., $\ell(f(x),\bar{y})=-\sum_{i=1}^{C}1_{\{\bar{y}=i\}}\log(g_i(x))$.
Let $\hat{f}$ and $\Delta \hat{T}$ be the learned classifier and slack variable.
Assume $\Delta \hat{T}$ is searched from a space of $\Delta T$ constituting valid transition matrices\footnote{During the training, $T+\Delta T$ can be ensured to be a valid transition matrix by first projecting their negative entries to be zero and then performing row normalization. In the experiments, $\Delta T$ is initialized to be a zero matrix and we haven't pushed $T+\Delta T$ to be a valid matrix when tuning $\Delta T$.},
i.e., $\forall \Delta T$ and $\forall i\neq j$, $\hat{T}_{ij}+\Delta {T}_{ij}\geq0$ and $\hat{T}_{ii}+\Delta {T}_{ii}>\hat{T}_{ij}+\Delta {T}_{ij}$.
Then, for any $\delta>0$, with probability at least $1-\delta$,
\begin{align}
\label{eq:theorem}%
\mathbb{E}[\bar{R}_{n,w}(\hat{T}+\Delta \hat{T},\hat{f})]- \bar{R}_{n,w}(\hat{T}+\Delta \hat{T},\hat{f})\leq \frac{2BCL(\sqrt{2d\log2}+1)\Pi_{i=1}^{d}M_i}{\sqrt{n}}+CM\sqrt{\frac{\log{1/\delta}}{2n}}.\nonumber
\end{align}
\end{theorem}
A detailed proof is provided in Appendix C. The factor $(\sqrt{2d\log2}+1)\Pi_{i=1}^{d}M_i$ is induced by the hypothesis complexity of the deep neural network \cite{golowich2018size} (see Theorem 1 therein), which could be improved \cite{neyshabur2017exploring,zhang2017understanding,kawaguchi2017generalization}. Although the proposed reweighted loss is more complex than the traditional unweighted loss function, we have derived a generalization error bound not larger than those derived for the algorithms employing the traditional loss \cite{mohri2018foundations} (can be seen by Lemma 2 in the proof of the theorem). This shows that the proposed Algorithm \ref{alg:reweighting} does not need a larger training sample to achieve a small difference between training error ($\bar{R}_{n,w}(\hat{T}+\Delta \hat{T},\hat{f})$) and test error ($\mathbb{E}[\bar{R}_{n,w}(\hat{T}+\Delta \hat{T},\hat{f})]$). Also note that deep learning is powerful in yielding a small training error. If the training sample size $n$ is large, then the upper bound in Theorem \ref{thm:main} is small, which implies a small $\mathbb{E}[\bar{R}_{n,w}(\hat{T}+\Delta \hat{T},\hat{f})]$ and justifies why the proposed method will have small test errors in the experiment section. Meanwhile, in the experiment section, we show that the proposed method is much superior to the state-of-the-art methods in classification accuracy, implying that the small generalization error is not obtained at the cost of enlarging the approximation error.

\section{Experiments}\label{sec:experiment}

\textbf{Datasets}\ \ We verify the effectiveness of the proposed method on three synthetic noisy datasets, i.e., \textit{MNIST} \cite{LeCunmnist}, \textit{CIFAR-10} \cite{krizhevsky2009learning}, and \textit{CIFAR-100} \cite{krizhevsky2009learning}, and one real-world noisy dataset, i.e., \textit{clothing1M} \cite{xiao2015learning}. 
\textit{MNIST} has 10 classes of images including 60,000 training images and 10,000 test images.
\textit{CIFAR-10} has 10 classes of images including 50,000 training images and 10,000 test images. \textit{CIFAR-100} also has 50,000 training images and 10,000 test images, but 100 classes. For all the datasets, we leave out 10\% of the training examples as a validation set. The three datasets contain clean data. We corrupted the training and validation sets manually according to true transition matrices $T$. Specifically, we employ the symmetry flipping setting defined in Appendix D. Sym-50 generates heavy label noise and leads almost half of the instances to have noisy labels, while Sym-20 generates light label noise and leads around 20\% of instances to have label noise. Note that the pair flipping setting \cite{han2018co}, where each row of the transition matrix only has two non-zero entries, has also been widely studied. However, for simplicity, we do not pose any constraint on the slack variable $\Delta T$ to achieve specific speculation of the transition matrix, e.g., sparsity \cite{han2018masking}. We leave this for future work.

Besides reporting the classification accuracy on test set, we also report the discrepancy between the learned transition matrix $\hat{T}+\Delta\hat{T}$ and the true one $T$. All experiments are repeated five times on those three datasets. \textit{Clothing1M} consists of 1M images with real-world noisy labels, and additional 50k, 14k, 10k images with clean labels for training, validation, and testing. We use the 50k clean data to help initialize the transition matrix as did in the baseline \cite{patrini2017making}.

\begin{table}[t]
	\centering
	\caption{Means and standard deviations (percentage) of classification accuracy. Methods with ``-A'' means that they run on the intact datasets without removing possible anchor points; Methods with ``-R'' means that the transition matrix used is revised by a revision $\Delta \hat{T}$.}
	\label{tab:accu2}
	\scalebox{0.8}
	{
		\begin{tabular}{c  c  c  c  c  c c}
		\toprule
% 		\cline{1-7}
		\multirow{2}*{}	&\multicolumn{2}{c}{\textit{MNIST}} & \multicolumn{2}{c}{\textit{CIFAR-10}} & \multicolumn{2}{c}{\textit{CIFAR-100}} \\
			~  & Sym-20\%      & Sym-50\%      & Sym-20\%       & Sym-50\%    & Sym-20\%  & Sym-50\%  \\ \midrule
			 Decoupling-A & 95.39$\pm$0.29 &81.52$\pm$0.29 & 79.85$\pm$0.30 & 52.22$\pm$0.45 &42.75$\pm$0.49        &29.24$\pm$0.54 \\
			 MentorNet-A & 96.57$\pm$0.18 & 90.13$\pm$0.09 & 80.49$\pm$0.52 & 70.71$\pm$0.24 & 52.11$\pm$0.10         & 38.45$\pm$0.25\\
			 Co-teaching-A & 97.22$\pm$0.18 & 91.68$\pm$0.21 & 82.38$\pm$0.11 & 72.80$\pm$0.45 & 54.23$\pm$0.08         & 41.37$\pm$0.08 \\
			  Forward-A & 98.75$\pm$0.08 & 97.86$\pm$0.22 & 85.63$\pm$0.52 & 77.92$\pm$0.66 & 57.75$\pm$0.37         & 44.66$\pm$1.01\\
		     Reweight-A& 98.71$\pm$0.11 & 98.13$\pm$0.19 & 86.77$\pm$0.40 & 80.16$\pm$0.46          & 58.35$\pm$0.64 &43.97$\pm$0.67\\ \midrule
			 Forward-A-R  & 98.84$\pm$0.09 & 98.12$\pm$0.22 & 88.10$\pm$0.21 & 81.11$\pm$0.74 & 62.13$\pm$2.09        & \textbf{50.46$\pm$0.52} \\
			 Reweight-A-R & \textbf{98.91$\pm$0.04} & \textbf{98.38$\pm$0.21} & \textbf{89.63$\pm$0.13} & \textbf{83.40$\pm$0.65} & \textbf{65.40$\pm$1.07}        & 50.24$\pm$1.45\\
			 \toprule
		\end{tabular}
	}
\end{table}

\begin{table}[t]
	\centering
	\caption{Means and standard deviations (percentage) of classification accuracy. Methods with ``-N/A'' means instances with high estimated $P({Y}|X)$ are removed from the dataset; Methods with ``-R'' means that the transition matrix used is revised by a revision $\Delta \hat{T}$.}
	\label{tab:accu3}
	\scalebox{0.8}
	{
		\begin{tabular}{c  c  c  c  c  c c}
		\toprule
% 		\cline{1-7}
		\multirow{2}*{}	&\multicolumn{2}{c}{\textit{MNIST}} & \multicolumn{2}{c}{\textit{CIFAR-10}} & \multicolumn{2}{c}{\textit{CIFAR-100}} \\
			~  & Sym-20\%      & Sym-50\%      & Sym-20\%       & Sym-50\%    & Sym-20\%  & Sym-50\%  \\ \midrule
		    Decoupling-N/A &95.93$\pm$0.21 &82.55$\pm$0.39 & 75.37$\pm$1.24 & 47.19$\pm$0.19 & 39.59$\pm$0.42         & 24.04$\pm$1.19 \\
			 MentorNet-N/A & $97.11\pm$0.09 & 91.44$\pm$0.25 & 78.51$\pm$0.31 & 67.37$\pm$0.30 & 48.62$\pm$0.43         & 33.53$\pm$0.31 \\
			 Co-teaching-N/A & 97.69$\pm$0.23 & 93.58$\pm$0.49 &81.72$\pm$0.14 &70.44$\pm$1.01 & 53.21$\pm$0.54         &40.06$\pm$0.83 \\
		     Forward-N/A& 98.64$\pm$0.12 & 97.74$\pm$0.13 & 84.75$\pm$0.81 & 74.32$\pm$0.69          & 56.23$\pm$0.34 &39.28$\pm$0.59\\
		     Reweight-N/A & 98.69$\pm$0.08 & 98.05$\pm$0.22 & 85.53$\pm$0.26 & 77.70$\pm$1.00 & 56.60$\pm$0.71         & 39.28$\pm$0.71\\ 
		     \midrule
		     Forward-N/A-R& 98.80$\pm$0.06 & 97.96$\pm$0.13 & 86.93$\pm$0.39 & 77.14$\pm$0.65          & 58.72$\pm$0.45 &44.60$\pm$0.79\\
		     Reweight-N/A-R & \textbf{98.85$\pm$0.02} & \textbf{98.37$\pm$0.17} & \textbf{88.90$\pm$0.22} & \textbf{81.55$\pm$0.94} & \textbf{62.00$\pm$1.78}         & \textbf{44.75$\pm$2.10}\\ 			  
		     \bottomrule
		\end{tabular}
	}
\end{table}

\textbf{Network structure and optimization}\ \ 
For fair comparison, we implement all methods with default parameters by PyTorch on NVIDIA Tesla V100. We use a LeNet-5 network for \textit{MNIST}, a ResNet-18 network for \textit{CIFAR-10}, a ResNet-34 network for \textit{CIFAR-100}. 
For learning the transition matrix $\hat{T}$ in the first stage, we follow the optimization method in \cite{patrini2017making}.
During the second stage, we first use SGD with momentum 0.9, weight decay $10^{-4}$, batch size 128, and an initial learning rate of $10^{-2}$ to initialize the network. The learning rate is divided by 10 after the 40th epoch and 80th epoch. 200 epochs are set in total. Then, the optimizer and learning rate are changed to Adam and $5\times 10^{-7}$ to learn the classifier and slack variable. For \textit{CIFAR-10} and \textit{CIFAR-100}, we perform data augmentation by horizontal random flips and 32$\times$32 random crops after padding 4 pixels on each side.
For \textit{clothing1M}, we use a ResNet-50 pre-trained on ImageNet. Follow \cite{patrini2017making}, we also exploit the 1M noisy data and 50k clean data to initialize the transition matrix. In the second stage, for initialization, we use SGD with momentum 0.9, weight decay $10^{-3}$, batch size 32, and run with learning rates $10^{-3}$ and $10^{-4}$ for 5 epochs each. For learning the classifier and slack variable, Adam is used and the learning rate is changed to $5\times 10^{-7}$.

\textbf{Baselines}\ \
We compare the proposed method with state-of-the-art approaches. Specifically, we compare with the following three inconsistent but well-designed algorithms: Decoupling \cite{malach2017decoupling}, MentorNet \cite{jiang2018mentornet}, and Co-teaching \cite{han2018co}, which free the learning of transition matrices. To compare with consistent estimators, we set Forward \cite{patrini2017making}, a classifier-consistent algorithm, and the importance reweighting method (Reweight), a risk-consistent algorithm, as baselines. The risk-consistent estimator involving the inverse of transition matrix, e.g., Backward in \cite{patrini2017making}, has not been included in the comparison, because it has been reported to perform worse than the Forward method \cite{patrini2017making}.

\subsection{Comparison for classification accuracy}

\textbf{The importance of anchor points}\ \
To show the importance of anchor points, we modify the datasets by moving possible anchor points, i.e., instances with large estimated class posterior probability $P({Y}|X)$, before corrupting the training and validation sets. As the \textit{MNIST} dataset is simple, we removed 40\% of the instances with the largest estimated class posterior probabilities in each class. For \textit{CIFAR-10} and \textit{CIFAR-100}, we removed 20\% of the instances with the largest estimated class posterior probabilities in each class.  To make it easy for distinguishing, we mark a ``-A'' in the algorithm's name if it runs on the original intact datasets, and mark a ``-N/A'' in the algorithm's name if it runs on those modified datasets.

Comparing Decoupling-A, MentorNet-A, and Co-teaching-A in Table \ref{tab:accu2} with Decoupling-N/A, MentorNet-N/A, and Co-teaching-N/A in Table \ref{tab:accu3}, we can find that on \textit{MNIST}, the methods with ``-N/A'' work better; while on \textit{CIFAR-10} and \textit{CIFAR-100}, the methods with ``-A'' work better. This is because those methods are independent of transition matrices but dependent of dataset properties. Removing possible anchors points may not always lead to performance degeneration.

Comparing Forward-A and Reweight-A with Forward-N/A and Reweight-N/A, we can find that the methods without anchor points, i.e., with ``-N/A'', degenerate clearly. The degeneration on \textit{MNIST} is slight because the dataset can be well separated and many instances have high class posterior probability even in the modify dataset. Those results show that, without anchor points, the consistent algorithms will have performance degeneration.
Specifically, on \textit{CIFAR-100}, the methods with ``-N/A'' have much worse performance than the ones with ``-A'', with accuracy dropping at least 4\%.

\begin{table}[t]
	\centering
	\caption{Means and standard deviations (percentage) of classification accuracy on \textit{MNIST} with different label noise levels. Methods with ``-A'' means that they run on the intact datasets without removing possible anchor points; Methods with ``-R'' means that the transition matrix used is revised by a revision $\Delta \hat{T}$; Methods with ``-N/A'' means instances with high estimated $P({Y}|X)$ are removed from the dataset. }
	\label{tab:accu5}
	\scalebox{0.8}
	{
		\begin{tabular}{cccc}
		%\hline\cline{1-5}
           	\toprule
			 ~& Sym-60\%      & Sym-70\%      & Sym-80\%            \\ \midrule
			 Forward-A & 97.10$\pm$0.08 & 96.06$\pm$0.41 & 91.46$\pm$1.03 \\
             Forward-A-R  & 97.65$\pm$0.11 & 96.42$\pm$0.35 & 91.77$\pm$0.22 \\ 
		     Reweight-A& 97.39$\pm$0.27 & 96.25$\pm$0.26 & 93.79$\pm$0.52  \\
			 Reweight-A-R & \textbf{97.83$\pm$0.18} & \textbf{97.13$\pm$0.08} & \textbf{94.19$\pm$0.45}  \\\midrule
			  Forward-N/A & 96.82$\pm$0.14 & 94.61$\pm$0.28 & 85.95$\pm$1.01 \\
			   Forward-N/A-R  & 96.99$\pm$0.16 & 95.02$\pm$0.17 & 86.04$\pm$1.03 \\
			   Reweight-N/A& 97.01$\pm$0.20 & 95.94$\pm$0.14 & 91.59$\pm$0.70 \\
			   Reweight-N/A-R & \textbf{97.81$\pm$0.12} & \textbf{96.59$\pm$0.15} & \textbf{91.91$\pm$0.65}\\
             \bottomrule
		\end{tabular}
	}
\end{table}

\begin{table}[t]
	\centering
	\caption{Classification accuracy  (percentage) on \textit{Clothing1M}.}
	\label{tab:accu4}
	\scalebox{0.8}
	{
		\begin{tabular}{c  c  c  c  c  c c}
		\toprule
% 		\cline{1-7}
		     Decoupling & MentorNet & Co-teaching & Forward & Reweight & Forward-R         & Reweight-R \\  \midrule
			  53.98 & 56.77 & 58.68  & 71.79 & 70.95 & 72.25        & \textbf{74.18} \\
			  \bottomrule
		\end{tabular}
	}
\end{table}

To discuss the model performances on \textit{MNIST} with more label noise, we raise the noise rates to 60\%, 70\%, 80\%. Other experiment settings are unchanged. The results are presented in Table \ref{tab:accu5}. We can see that the proposed model outperforms the baselines more significantly as the noise rate grows. 

\textbf{Risk-consistent estimator vs.~classifier-consistent estimator}\ \ 
Comparing Forward-A with Reweight-A in Table \ref{tab:accu2} and comparing Forward-N/A with Reweight-N/A in Table \ref{tab:accu3}, it can be seen that the proposed Reweight method, a risk-consistent estimator not involving the inverse of transition matrix, works slightly better than or is comparable to Forward, a classifier-consistent algorithm. 
Note that in \cite{patrini2017making}, it is reported that Backward, a risk-consistent estimator which involves the inverse of the transition matrix, works worse than Forward, the classifier-consistent algorithm.

\begin{figure}[t]
\centering
\includegraphics[width=0.3\textwidth]{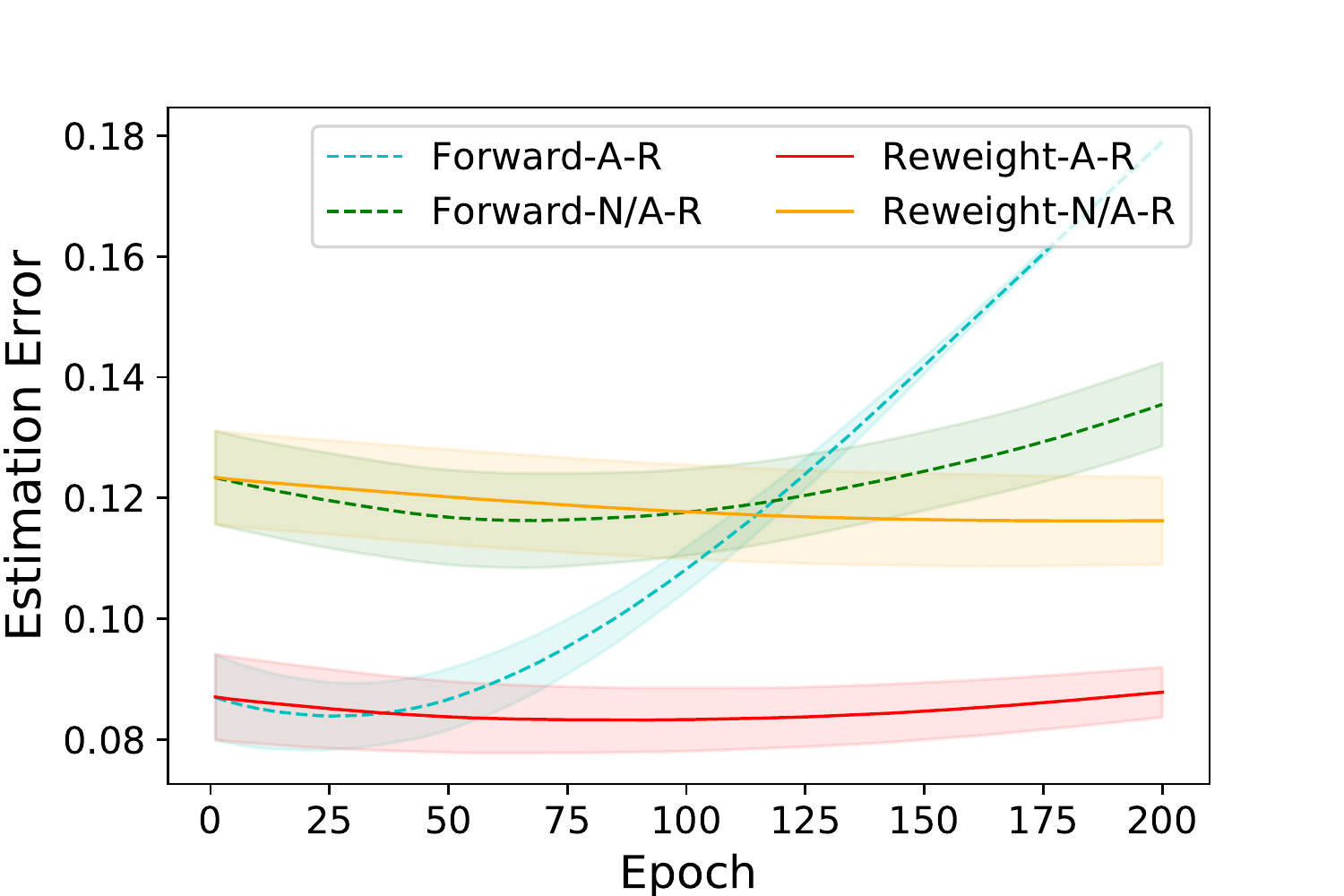}
\includegraphics[width=0.3\textwidth]{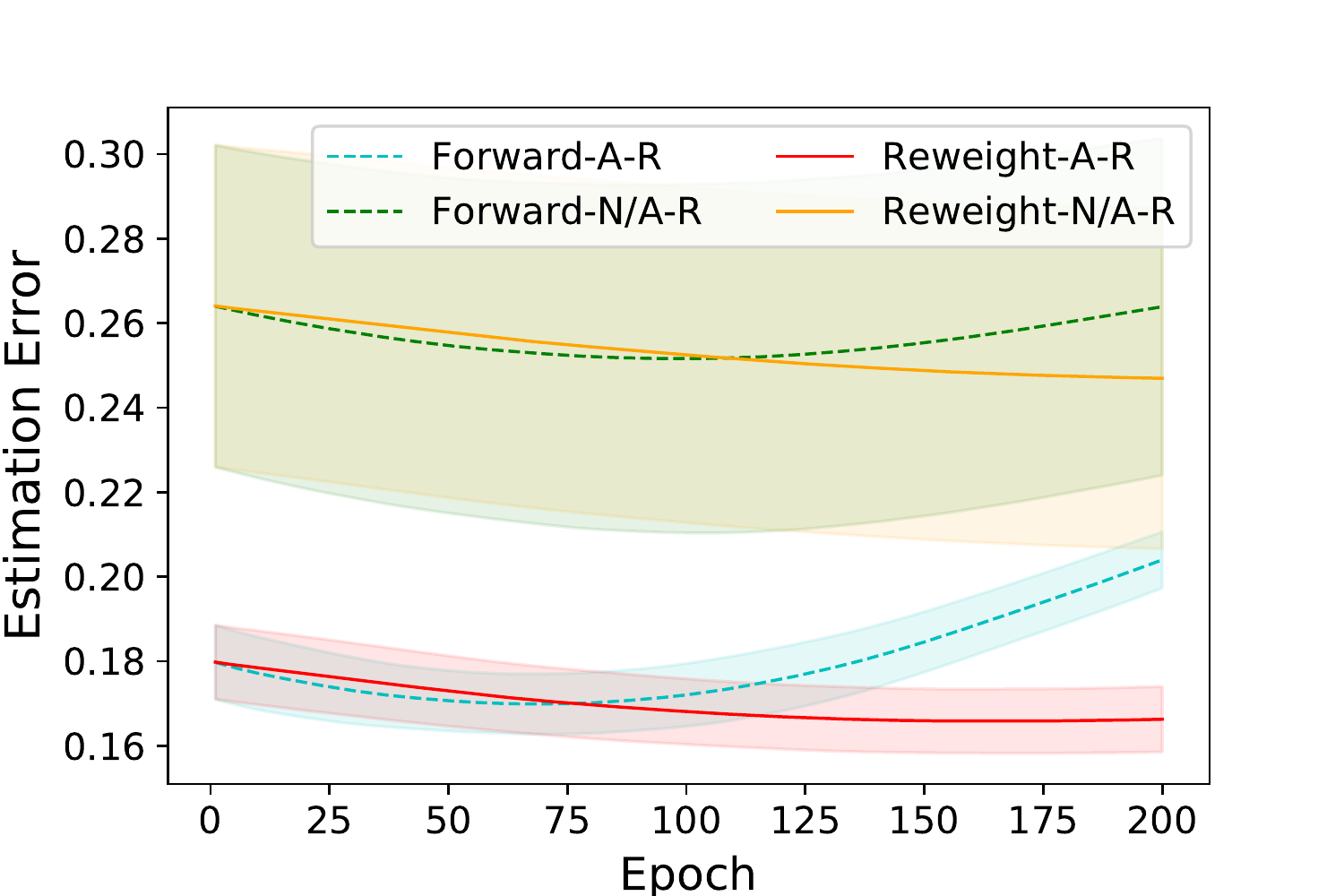}
\includegraphics[width=0.3\textwidth]{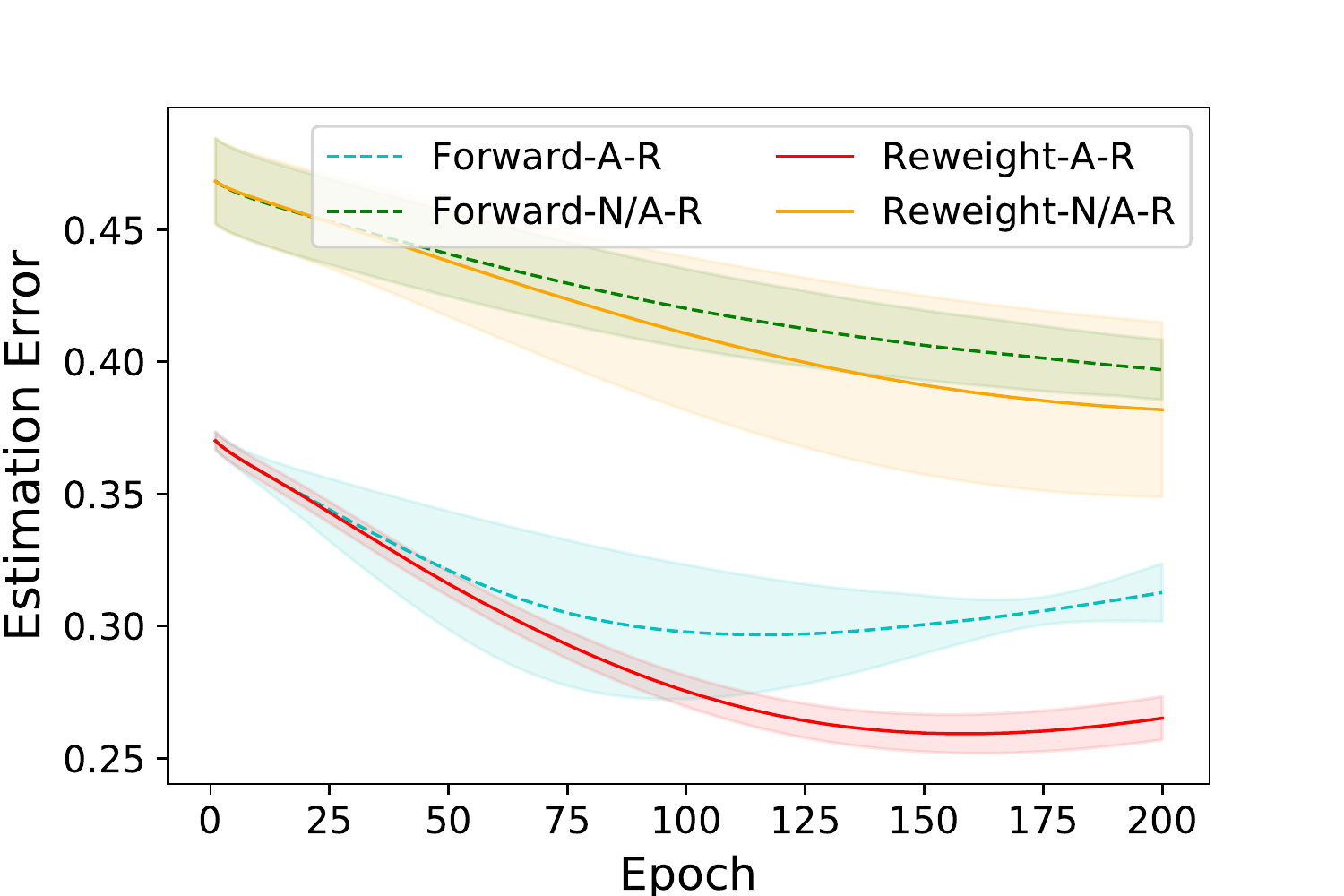}
\subfigure[\textit{MNIST}]
{\includegraphics[width=0.3\textwidth]{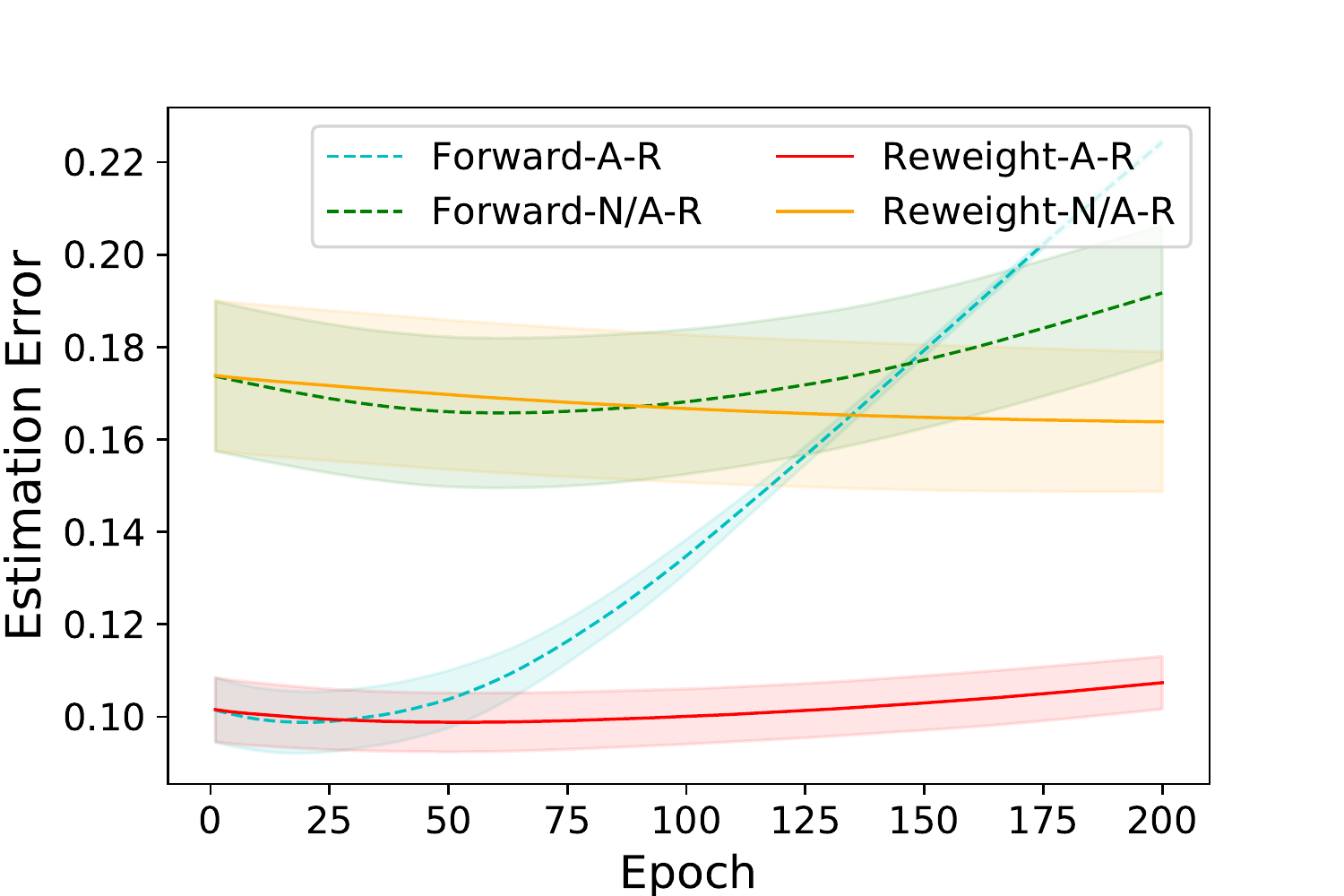}}
\subfigure[\textit{CIFAR-10}]
{\includegraphics[width=0.3\textwidth]{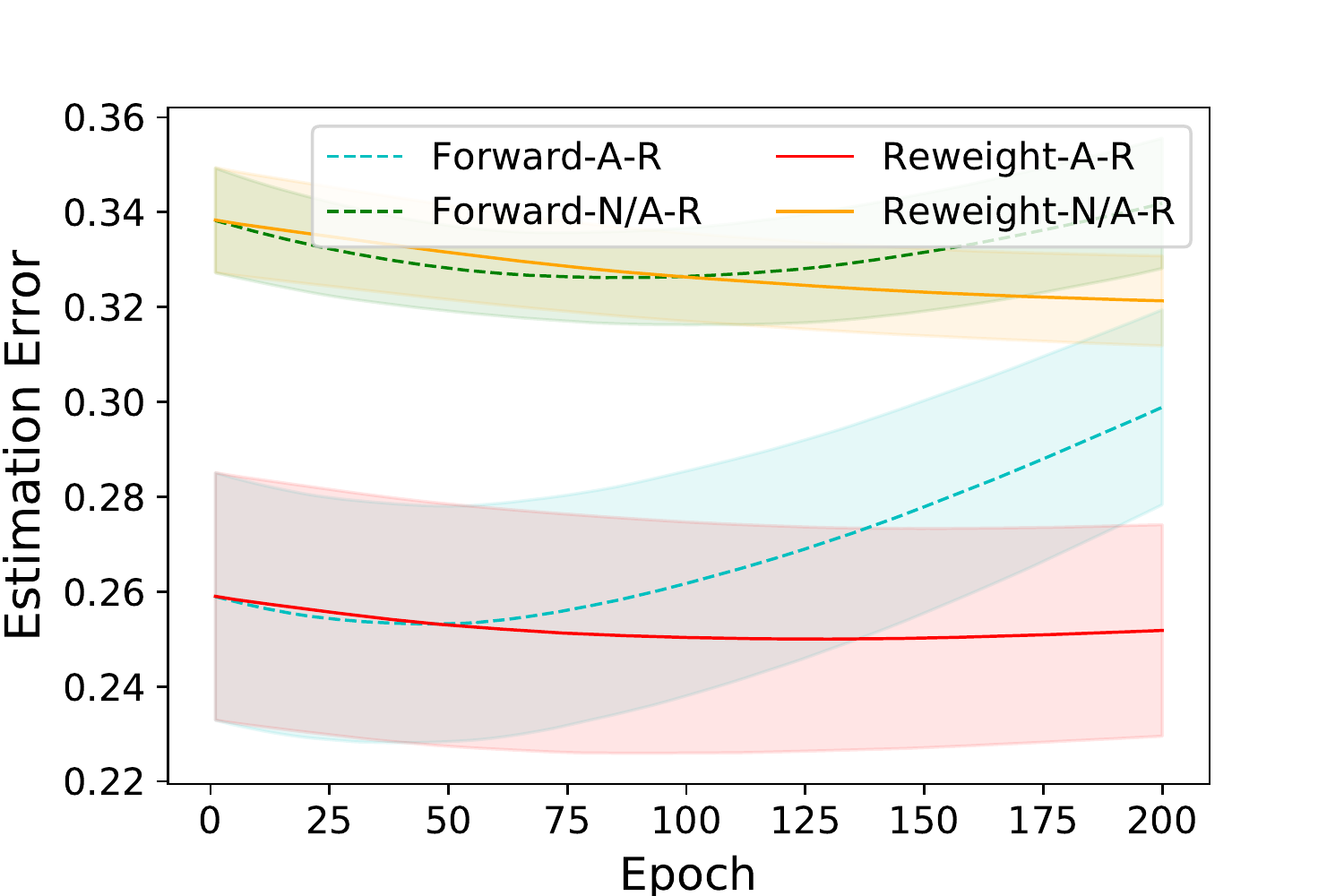}}
\subfigure[\textit{CIFAR-100}]
{\includegraphics[width=0.3\textwidth]{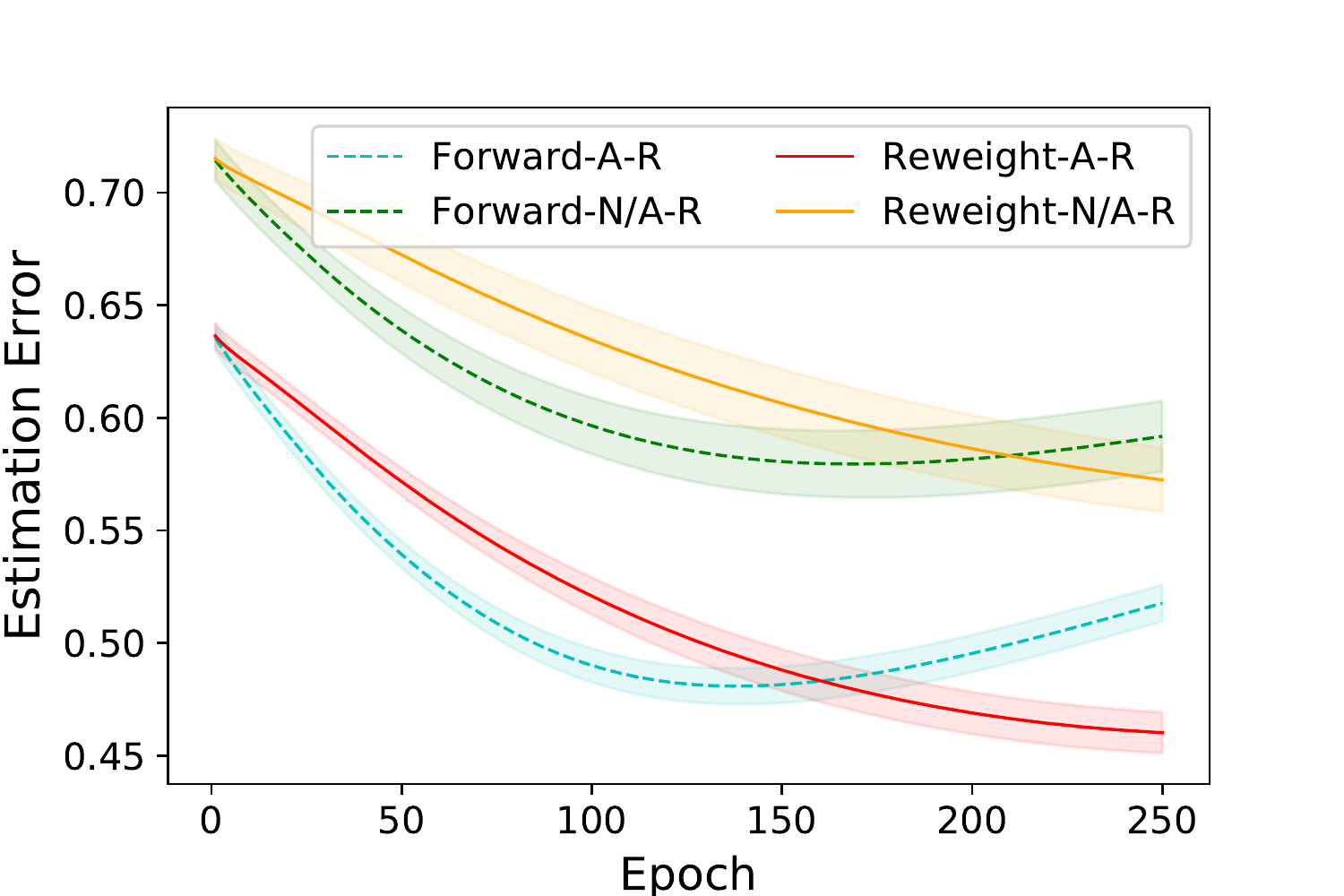}}
\caption{The estimation error of the transition matrix by employing classifier-consistent and risk-consistent estimators. The first row is about sym-20 label noise while the second row is about sym-50 label noise. The error bar for standard deviation in each figure has been shaded. }
\label{fig:estimation}
\end{figure}

\textbf{The importance of $T$-revision}\ \
Note that for fair comparison, we also set it as a baseline to modify the transition matrix in Forward.
As shown in Tables \ref{tab:accu2} and \ref{tab:accu3}, methods with ``-R'' means that they use the proposed $T$-revision method, i.e., modify the learned $\hat{T}$ by adding $\Delta \hat{T}$. Comparing the results in Tables \ref{tab:accu2} and \ref{tab:accu3}, {we can find that the $T$-revision method significantly outperforms the others.} Among them, the proposed Reweight-R works significantly better than the baseline Forward-R. We can find that the $T$-Revision method boosts the classification performance even without removing possible anchor points.
The rationale behind this may be that the network, transition matrix, and classifier are jointly learned and validated and that the identified anchor points are not reliable.

\textbf{Comparison on real-world dataset}\ \
The proposed $T$-revision method significantly outperforms the baselines as shown in Table \ref{tab:accu4}, where the highest accuracy is bold faced.

\subsection{Comparison for estimating transition matrices}
To show that the proposed risk-consistent estimator is more effective in modifying the transition matrix, we plot the estimation error for the transition matrix, i.e., $\|T-\hat{T}-\Delta \hat{T}\|_1/\|T\|_1$. 
% Note that the $Q$ revision method is not theoretically guaranteed to learn a better transition matrix.
In Figure \ref{fig:estimation}, we can see that for all cases, the proposed risk-consistent-estimator-based revision leads to smaller estimator errors than the classifier-consistent algorithm based method (Forward-R), showing that the risk-consistent estimator is more powerful in modifying the transition matrix. This also explains why the proposed method works better. We provide more discussions about Figure \ref{fig:estimation} in Appendix E.

\section{Conclusion}\label{sec:conclusion}
This paper presents a risk-consistent estimator for label-noise learning without involving the inverse of transition matrix and a simple but effective learning paradigm called $T$-revision, which trains
deep neural networks robustly under noisy supervision. The aim is to maintain effectiveness and efficiency of current consistent algorithms when there are no anchor points and then the transition matrices are poorly learned. The key idea is to revise the learned transition matrix and validate the revision by exploiting a noisy validation set. We conduct experiments on both synthetic and real-world label noise data to demonstrate that the proposed $T$-revision can significantly help boost the performance of label-noise learning. In the future, we will extend the work in the following aspects. First, how to incorporate some prior knowledge of the transition matrix, e.g., sparsity, into the end-to-end learning system. Second, how to recursively learn the transition matrix and classifier as our experiments show that transition matrices can be refined.

\section*{Acknowledgments}
TLL was supported by Australian Research Council Project DP180103424 and DE190101473.
NNW was supported by National Natural Science Foundation
of China under Grants  61922066, 61876142, and the CCF-Tencent Open Fund.
CG was supported by NSF of China under Grants 61602246, 61973162, NSF of Jiangsu Province under Grants BK20171430, the Fundamental Research Funds for the Central Universities under Grants 30918011319, and the “Young Elite Scientists Sponsorship Program” by CAST under Grants 2018QNRC001.
MS was supported by the International Research Center for Neurointelligence (WPI-IRCN) at The University of Tokyo Institutes for Advanced Study.
XBX and TLL would give special thanks to Haifeng Liu and Brain-Inspired Technology Co., Ltd. for their support of GPUs used for this research.

\bibliographystyle{plain}
\bibliography{bib}

\begin{thebibliography}{10}

\bibitem{angluin1988learning}
Dana Angluin and Philip Laird.
\newblock Learning from noisy examples.
\newblock {\em Machine Learning}, 2(4):343--370, 1988.

\bibitem{bartlett2017spectrally}
Peter~L Bartlett, Dylan~J Foster, and Matus~J Telgarsky.
\newblock Spectrally-normalized margin bounds for neural networks.
\newblock In {\em NeurIPS}, pages 6240--6249, 2017.

\bibitem{bartlett2006convexity}
Peter~L Bartlett, Michael~I Jordan, and Jon~D McAuliffe.
\newblock Convexity, classification, and risk bounds.
\newblock {\em Journal of the American Statistical Association},
  101(473):138--156, 2006.

\bibitem{bartlett2002rademacher}
Peter~L Bartlett and Shahar Mendelson.
\newblock Rademacher and gaussian complexities: Risk bounds and structural
  results.
\newblock {\em Journal of Machine Learning Research}, 3(Nov):463--482, 2002.

\bibitem{blanchard2010semi}
Gilles Blanchard, Gyemin Lee, and Clayton Scott.
\newblock Semi-supervised novelty detection.
\newblock {\em Journal of Machine Learning Research}, 11(Nov):2973--3009, 2010.

\bibitem{boucheron2005theory}
St{\'e}phane Boucheron, Olivier Bousquet, and G{\'a}bor Lugosi.
\newblock Theory of classification: A survey of some recent advances.
\newblock {\em ESAIM: probability and statistics}, 9:323--375, 2005.

\bibitem{boucheron2013concentration}
St{\'e}phane Boucheron, G{\'a}bor Lugosi, and Pascal Massart.
\newblock {\em Concentration inequalities: A nonasymptotic theory of
  independence}.
\newblock Oxford university press, 2013.

\bibitem{cheng2017learning}
Jiacheng Cheng, Tongliang Liu, Kotagiri Ramamohanarao, and Dacheng Tao.
\newblock Learning with bounded instance-and label-dependent label noise.
\newblock {\em arXiv preprint arXiv:1709.03768}, 2017.

\bibitem{goldberger2016training}
Jacob Goldberger and Ehud Ben-Reuven.
\newblock Training deep neural-networks using a noise adaptation layer.
\newblock In {\em ICLR}, 2017.

\bibitem{golowich2018size}
Noah Golowich, Alexander Rakhlin, and Ohad Shamir.
\newblock Size-independent sample complexity of neural networks.
\newblock In {\em COLT}, pages 297--299, 2018.

\bibitem{gretton2009covariate}
Arthur Gretton, Alex Smola, Jiayuan Huang, Marcel Schmittfull, Karsten
  Borgwardt, and Bernhard Sch{\"o}lkopf.
\newblock Covariate shift by kernel mean matching.
\newblock {\em Dataset shift in machine learning}, pages 131--160, 2009.

\bibitem{guo2018curriculumnet}
Sheng Guo, Weilin Huang, Haozhi Zhang, Chenfan Zhuang, Dengke Dong, Matthew~R
  Scott, and Dinglong Huang.
\newblock Curriculumnet: Weakly supervised learning from large-scale web
  images.
\newblock In {\em ECCV}, pages 135--150, 2018.

\bibitem{han2018masking}
Bo~Han, Jiangchao Yao, Gang Niu, Mingyuan Zhou, Ivor Tsang, Ya~Zhang, and
  Masashi Sugiyama.
\newblock Masking: A new perspective of noisy supervision.
\newblock In {\em NeurIPS}, pages 5836--5846, 2018.

\bibitem{han2018co}
Bo~Han, Quanming Yao, Xingrui Yu, Gang Niu, Miao Xu, Weihua Hu, Ivor Tsang, and
  Masashi Sugiyama.
\newblock Co-teaching: Robust training of deep neural networks with extremely
  noisy labels.
\newblock In {\em NeurIPS}, pages 8527--8537, 2018.

\bibitem{jiang2018mentornet}
Lu~Jiang, Zhengyuan Zhou, Thomas Leung, Li-Jia Li, and Li~Fei-Fei.
\newblock {MentorNet}: Learning data-driven curriculum for very deep neural
  networks on corrupted labels.
\newblock In {\em ICML}, pages 2309--2318, 2018.

\bibitem{kawaguchi2017generalization}
Kenji Kawaguchi, Leslie~Pack Kaelbling, and Yoshua Bengio.
\newblock Generalization in deep learning.
\newblock {\em arXiv preprint arXiv:1710.05468}, 2017.

\bibitem{kremer2018robust}
Jan Kremer, Fei Sha, and Christian Igel.
\newblock Robust active label correction.
\newblock In {\em AISTATS}, pages 308--316, 2018.

\bibitem{krizhevsky2009learning}
Alex Krizhevsky.
\newblock Learning multiple layers of features from tiny images.
\newblock Technical report, 2009.

\bibitem{LeCunmnist}
Yann LeCun, Corinna Cortes, and Christopher~J.C. Burges.
\newblock The {MNIST} database of handwritten digits.
\newblock {\em http://yann.lecun.com/exdb/mnist/}.

\bibitem{ledoux2013probability}
Michel Ledoux and Michel Talagrand.
\newblock {\em Probability in Banach Spaces: isoperimetry and processes}.
\newblock Springer Science \& Business Media, 2013.

\bibitem{li2017learning}
Yuncheng Li, Jianchao Yang, Yale Song, Liangliang Cao, Jiebo Luo, and Li-Jia
  Li.
\newblock Learning from noisy labels with distillation.
\newblock In {\em ICCV}, pages 1910--1918, 2017.

\bibitem{liu2016classification}
Tongliang Liu and Dacheng Tao.
\newblock Classification with noisy labels by importance reweighting.
\newblock {\em IEEE Transactions on pattern analysis and machine intelligence},
  38(3):447--461, 2016.

\bibitem{ma2018dimensionality}
Xingjun Ma, Yisen Wang, Michael~E Houle, Shuo Zhou, Sarah~M Erfani, Shu-Tao
  Xia, Sudanthi Wijewickrema, and James Bailey.
\newblock Dimensionality-driven learning with noisy labels.
\newblock In {\em ICML}, pages 3361--3370, 2018.

\bibitem{malach2017decoupling}
Eran Malach and Shai Shalev-Shwartz.
\newblock Decoupling" when to update" from" how to update".
\newblock In {\em NeurIPS}, pages 960--970, 2017.

\bibitem{mohri2018foundations}
Mehryar Mohri, Afshin Rostamizadeh, and Ameet Talwalkar.
\newblock {\em Foundations of Machine Learning}.
\newblock MIT Press, 2018.

\bibitem{natarajan2013learning}
Nagarajan Natarajan, Inderjit~S Dhillon, Pradeep~K Ravikumar, and Ambuj Tewari.
\newblock Learning with noisy labels.
\newblock In {\em NeurIPS}, pages 1196--1204, 2013.

\bibitem{neyshabur2017exploring}
Behnam Neyshabur, Srinadh Bhojanapalli, David McAllester, and Nati Srebro.
\newblock Exploring generalization in deep learning.
\newblock In {\em NeurIPS}, pages 5947--5956, 2017.

\bibitem{neyshabur2018pac}
Behnam Neyshabur, Srinadh Bhojanapalli, and Nathan Srebro.
\newblock A {PAC}-{B}ayesian approach to spectrally-normalized margin bounds
  for neural networks.
\newblock In {\em ICLR}, 2018.

\bibitem{northcuttlearning}
Curtis~G Northcutt, Tailin Wu, and Isaac~L Chuang.
\newblock Learning with confident examples: Rank pruning for robust
  classification with noisy labels.
\newblock In {\em UAI}, 2017.

\bibitem{patrini2017making}
Giorgio Patrini, Alessandro Rozza, Aditya Krishna~Menon, Richard Nock, and
  Lizhen Qu.
\newblock Making deep neural networks robust to label noise: A loss correction
  approach.
\newblock In {\em CVPR}, pages 1944--1952, 2017.

\bibitem{ramaswamy2016mixture}
Harish Ramaswamy, Clayton Scott, and Ambuj Tewari.
\newblock Mixture proportion estimation via kernel embeddings of distributions.
\newblock In {\em ICML}, pages 2052--2060, 2016.

\bibitem{reed2014training}
Scott~E Reed, Honglak Lee, Dragomir Anguelov, Christian Szegedy, Dumitru Erhan,
  and Andrew Rabinovich.
\newblock Training deep neural networks on noisy labels with bootstrapping.
\newblock In {\em ICLR}, 2015.

\bibitem{ren2018learning}
Mengye Ren, Wenyuan Zeng, Bin Yang, and Raquel Urtasun.
\newblock Learning to reweight examples for robust deep learning.
\newblock In {\em ICML}, pages 4331--4340, 2018.

\bibitem{scott2012calibrated}
Clayton Scott.
\newblock Calibrated asymmetric surrogate losses.
\newblock {\em Electronic Journal of Statistics}, 6:958--992, 2012.

\bibitem{scott2015rate}
Clayton Scott.
\newblock A rate of convergence for mixture proportion estimation, with
  application to learning from noisy labels.
\newblock In {\em AISTATS}, pages 838--846, 2015.

\bibitem{scott2013classification}
Clayton Scott, Gilles Blanchard, and Gregory Handy.
\newblock Classification with asymmetric label noise: Consistency and maximal
  denoising.
\newblock In {\em COLT}, pages 489--511, 2013.

\bibitem{tanaka2018joint}
Daiki Tanaka, Daiki Ikami, Toshihiko Yamasaki, and Kiyoharu Aizawa.
\newblock Joint optimization framework for learning with noisy labels.
\newblock In {\em CVPR}, pages 5552--5560, 2018.

\bibitem{thekumparampil2018robustness}
Kiran~K Thekumparampil, Ashish Khetan, Zinan Lin, and Sewoong Oh.
\newblock Robustness of conditional gans to noisy labels.
\newblock In {\em NeurIPS}, pages 10271--10282, 2018.

\bibitem{vahdat2017toward}
Arash Vahdat.
\newblock Toward robustness against label noise in training deep discriminative
  neural networks.
\newblock In {\em NeurIPS}, pages 5596--5605, 2017.

\bibitem{vandermeulen2016operator}
Robert~A Vandermeulen and Clayton~D Scott.
\newblock An operator theoretic approach to nonparametric mixture models.
\newblock {\em arXiv preprint arXiv:1607.00071}, 2016.

\bibitem{vandermeulen2019operator}
Robert~A Vandermeulen and Clayton~D Scott.
\newblock An operator theoretic approach to nonparametric mixture models.
\newblock {\em accepted to The Annals of Statistics}, 2019.

\bibitem{vapnik2013nature}
Vladimir Vapnik.
\newblock {\em The nature of statistical learning theory}.
\newblock Springer science \& business media, 2013.

\bibitem{veit2017learning}
Andreas Veit, Neil Alldrin, Gal Chechik, Ivan Krasin, Abhinav Gupta, and Serge
  Belongie.
\newblock Learning from noisy large-scale datasets with minimal supervision.
\newblock In {\em CVPR}, pages 839--847, 2017.

\bibitem{xiao2015learning}
Tong Xiao, Tian Xia, Yi~Yang, Chang Huang, and Xiaogang Wang.
\newblock Learning from massive noisy labeled data for image classification.
\newblock In {\em CVPR}, pages 2691--2699, 2015.

\bibitem{yu2019does}
Xingrui Yu, Bo~Han, Jiangchao Yao, Gang Niu, Ivor~W Tsang, and Masashi
  Sugiyama.
\newblock How does disagreement benefit co-teaching?
\newblock In {\em ICML}, 2019.

\bibitem{yu2018efficient}
Xiyu Yu, Tongliang Liu, Mingming Gong, Kayhan Batmanghelich, and Dacheng Tao.
\newblock An efficient and provable approach for mixture proportion estimation
  using linear independence assumption.
\newblock In {\em CVPR}, pages 4480--4489, 2018.

\bibitem{yu2018learning}
Xiyu Yu, Tongliang Liu, Mingming Gong, and Dacheng Tao.
\newblock Learning with biased complementary labels.
\newblock In {\em ECCV}, pages 68--83, 2018.

\bibitem{zhang2017understanding}
Chiyuan Zhang, Samy Bengio, Moritz Hardt, Benjamin Recht, and Oriol Vinyals.
\newblock Understanding deep learning requires rethinking generalization.
\newblock In {\em ICLR}, 2017.

\bibitem{zhang2018generalized}
Zhilu Zhang and Mert Sabuncu.
\newblock Generalized cross entropy loss for training deep neural networks with
  noisy labels.
\newblock In {\em NeurIPS}, pages 8778--8788, 2018.

\end{thebibliography}

\newpage
\appendix
\section{Label-noise learning, or noisy-label learning, that is the point}

Note that the title of this paper is a question ``are anchor points really indispensable in \emph{label-noise learning}'' but not ``are anchor points really indispensable in \emph{noisy-label learning}''. Here, we explain why in order to make sense it must be label-noise learning. At first glance, label-noise learning may sound like we are learning the label noise, but this is exactly what we have implied in the title.

Generally speaking, the two names are synonyms of \emph{learning with noisy labels}---this is the title of \cite{natarajan2013learning} where the first statistically consistent learning method was proposed for training classifiers with noisy labels. For the family of consistent learning methods, the estimation of the transition matrix $T$ is always necessary. In fact, any such method requires three components:
\begin{itemize}
    \setlength\itemsep{0ex}\vspace{-1ex}%
    \item a label corruption process parameterized by $T$,
    \item an estimator of $T$, and
    \item a statistical or algorithmic correction using the estimated $T$.
    \vspace{-1ex}%
\end{itemize}
As a result, $T$ is also a target of learning, that is, the label noise is both a target to be learned and a source from which we learn. This is learning \emph{with} noisy labels, more than learning \emph{from} noisy labels.

For the first component, we assume the class-conditional noise model \cite{natarajan2013learning,patrini2017making}; for the third component, we rely on importance reweighting for learning with noisy labels \cite{liu2016classification}. Our novelty and major contribution is the second component, where we relax a requirement in existing consistent learning methods, namely we should have a certain amount of anchor points for estimating $T$ accurately, so that the correction using the estimated $T$ can be performed well.

While it is necessary to estimate $T$ for consistent learning methods, it is not the case for inconsistent learning methods. For instance, \emph{sample selection} methods try to remove mislabeled data, and \emph{label correction} methods try to fix the wrong labels of mislabeled data. None of them estimate $T$ so that none of them ever need the existence of anchor points.

Therefore, if we ask ``are anchor points really indispensable in label-noise learning'' where we are learning, modeling or estimating the label noise, the answer was \emph{yes} previously and is \emph{no} currently. Nevertheless, if we ask ``are anchor points really indispensable in noisy-label learning'' where some inconsistent learning method is employed without estimating the label noise, the answer has already been known to be \emph{no}. That is the point.

\section{How consistent algorithms work}
The aim of multi-class classification is to learn a \textit{hypothesis} $f$ that predicts labels for given instances. Typically, the hypothesis is of the following form: $f(x)=\arg\max_{i\in\{1,2,\ldots,C\}}g_i(x)$, where $g_i(x)$ is an estimate of $P(Y=i|X=x)$.
Let define the \textit{expected risk} of employing $f$ as 
\begin{eqnarray}
R(f)=\mathbb{E}_{(X,Y)\sim D}[\ell(f(X),Y)].
\end{eqnarray}
The \textit{optimal} hypothesis to learn is the one that minimizes the risk $R(f)$. 
Usually, the distribution $D$ is unknown. The optimal hypothesis is approximated by the minimizer of an \textit{empirical counterpart} of $R(f)$, i.e., the empirical risk
\begin{eqnarray}
R_n(f)=\frac{1}{n}\sum_{i=1}^{n}\ell(f(x_i),y_i).
\end{eqnarray}
The empirical risk $R_n(f)$ is \textbf{\textit{risk-consistent}} w.r.t. all loss functions, i.e., $R_n(f)\rightarrow R(f)$ as $n\rightarrow \infty$. Note that in the main paper, we have treated the training sample $\{X_i,\bar{Y}_i\}_{i=1}^n$ as iid variables to derive the generalization bound.

If the loss function is zero-one loss, i.e., $\ell(f(x),y)=1_{\{f(x)\neq y\}}$ where $1_{\{\cdot\}}$ is the indicator function and that the predefined \textit{hypothesis class} \cite{mohri2018foundations} is large enough, the optimal hypothesis that minimizing $R(f)$ is identical to the Bayes classifier \cite{bartlett2006convexity}, i.e.,
\begin{eqnarray}
f_\rho(x)=\arg\max_{i\in\{1,2,\ldots,C\}}P(Y=i|X=x).
\end{eqnarray}
Many frequently used loss functions are proven to be \textit{classification-calibrated} \cite{bartlett2006convexity,scott2012calibrated}, which means they will lead to classifiers having the same predictions as the classifier learned by using zero-one loss if the training sample size is sufficiently large  \cite{vapnik2013nature,mohri2018foundations}. In other words, the approximation, i.e., $\arg\min R_n(f)$, could converge to the optimal hypothesis by increasing the sample size $n$ and the corresponding estimator is therefore \textbf{\textit{classifier-consistent}}. Note that {risk-consistent} algorithm is also {classifier-consistent}. However, a classifier-consistent algorithm may not be risk-consistent.

% Let $P(Y|X) = [P(Y=1|X),\ldots,P(Y=C|X)]^\top$. According to the definition of transition matrix, we have that  $P(\bar{Y}|X)=T^\top P(Y|X)$, implying that if we let ${h}(X)=\arg\max_{i\in\{1,2,\ldots,C\}}(T^\top {g})_i(X)$, minimizing the empirical counterpart of $\bar{R}(h)=\mathbb{E}_{(X,\bar{Y})\sim \bar{D}}[\ell(h(X),\bar{Y})]$ by using only noisy data will lead to a \textbf{classifier-consistent estimator}, i.e., $\arg\max_{i\in\{1,2,\ldots,C\}} {g}_i(X)$. In other words, $\arg\max_{i\in\{1,2,\ldots,C\}} {g}_i(X)$ will converge to the optimal classifier for clean data by increasing the noisy sample size. That's why noise adaption layer has been widely used in deep learning to modify the softmax function (i.e., ${g}(X)$) \cite{goldberger2016training,patrini2017making,thekumparampil2018robustness,yu2018learning}.

% Note that if the loss function is zero-one loss, i.e., $\ell(f(X),Y)=1_{\{f(X)\neq Y\}}$ where $1_{\{\cdot\}}$ is the indicator function and that the predefined \textit{hypothesis class} \cite{mohri2018foundations} is large enough, the optimal hypothesis is identical to the Bayes classifier \cite{bartlett2006convexity}, i.e., $f_\rho(X)=\arg\max_{i\in\{1,2,\ldots,C\}}P(Y=i|X)$. Many frequently used loss functions are proven to be \textit{classification-calibrated} \cite{bartlett2006convexity,scott2012calibrated}, which means they will lead to classifiers having the same predictions as the classifier learned by using zero-one loss.

Given only the noisy training sample $\{(X_i,\bar{Y}_i)\}_{i=1}^{n}$, we have a noisy version of the empirical risk as 
\begin{eqnarray}
\bar{R}_n(f)=\frac{1}{n}\sum_{i=1}^{n}\ell({f}(X_i),\bar{Y}_i).
\end{eqnarray}
The learned ${{g}}(X)$ can be used to approximate $P({\bf \bar{Y}}|X)$. 
According to the definition of transition matrix, we have that  $P({\bf \bar{Y}}|X)=T^\top P({\bf {Y}}|X)$, implying that if we let 
\begin{eqnarray}
\bar{h}(X)=\arg\max_{i\in\{1,2,\ldots,C\}}(T^\top {g})_i(X),
\end{eqnarray}
minimizing 
\begin{eqnarray}
\bar{R}_n(\bar{h})=\frac{1}{n}\sum_{i=1}^{n}\ell(\bar{h}(X_i),\bar{Y}_i)
\end{eqnarray}
by using only noisy data will lead to a {classifier-consistent algorithm}. In other words, $\arg\max_{i\in\{1,2,\ldots,C\}} {g}_i(x)$ in the algorithm will converge to the optimal classifier for clean data by increasing the noisy sample size. That's why noise adaption layer has been widely used in deep learning to modify the softmax function (i.e., ${g}(x)$) \cite{goldberger2016training,patrini2017making,thekumparampil2018robustness,yu2018learning}.

If the transition matrix is invertable, the equation $P({\bf {Y}}|X)=(T^\top)^{-1} P({\bf \bar{Y}}|X)$ has been explored to design {risk-consistent estimator} for $R(f)$, e.g., \cite{natarajan2013learning,patrini2017making}. The basic idea is to modify the loss function $\ell({f}(X),\bar{Y})$ to be $\tilde{\ell}({f}(X),\bar{Y})$ such that for $X$ and $Y$, 
\begin{eqnarray}
\mathbb{E}_{\bar{Y}}[\tilde{\ell}({f}(X),\bar{Y})]=\ell(f(X),Y)
\end{eqnarray}
and thus 
\begin{eqnarray}
\mathbb{E}_{(X,Y,\bar{Y})}\tilde{\ell}({f}(X),\bar{Y})=R(f).
\end{eqnarray}
Specifically, 
let 
\begin{eqnarray}
{\mathcal{L}}(f(X),{\bf {Y}})=[{\ell}(f(X),{Y}=1),\ldots,{\ell}(f(X),{Y}=C)]^\top
\end{eqnarray}
and 
\begin{eqnarray}
\tilde{\mathcal{L}}(f(X),{\bf \bar{Y}})=[\tilde{\ell}(f(X),\bar{Y}=1),\ldots,\tilde{\ell}(f(X),\bar{Y}=C)]^\top=(T^\top)^{-1}{\mathcal{L}}(f(X),{\bf \bar{Y}}).
\end{eqnarray}
The losses $\tilde{\ell}$ will lead to risk-consistent estimator because  
\begin{eqnarray}
\mathbb{E}_{\bar{Y}|Y}[\tilde{\mathcal{L}}(f(X),{\bf \bar{Y}})]=T^\top\tilde{\mathcal{L}}(f(X),{\bf \bar{Y}})={\mathcal{L}}(f(X),{\bf {Y}}).
\end{eqnarray}
Risk-consistent algorithms are also classifier-consistent, but have some unique properties than classifier-consistent algorithms, e.g., can be used to tune hyper-parameter. {However, the current risk-consistent estimators contain the inverse of transition matrix, making parameter tuning inefficient and leading to performance degeneration. Our proposed risk-consistent estimator overcome the aforementioned issues.}

\section{Proof of Theorem 1}

We have defined
\begin{equation}
\bar{R}_{n,w}(\hat{T}+\Delta T,f)=\frac{1}{n}\sum_{i=1}^{n}\frac{{g}_{\bar{Y}_i}(X_i)}{((\hat{T}+\Delta T)^\top{g})_{\bar{Y}_i}(X_i)}\ell(f(X_i),\bar{Y}_i),
\end{equation}
where $f(X)=\arg\max_{i\in\{1,\ldots,C\}}{g}_i(X)$.
Let $S=\{(X_1,\bar{Y}_1),\ldots,(X_n,\bar{Y}_n)\}$, $S^i=\{(X_1,\bar{Y}_1),\ldots,(X_{i-1},\bar{Y}_{i-1}), (X'_i,\bar{Y}'_i), (X_{i+1},\bar{Y}_{i+1}), \ldots, (X_n,\bar{Y}_n)\}$, and
\begin{equation}
\Phi(S)=\sup_{\Delta T,f}(\bar{R}_{n,w}(\hat{T}+\Delta T,f)-\mathbb{E}_S[\bar{R}_{n,w}(\hat{T}+\Delta T,f)]).
\end{equation}

\begin{lemma}\label{lemma:1}
Let $\Delta \hat{T}$ and $\hat{f}$ be the learned slack variable and classifier respectively. Assume the learned transition matrix is valid,
i.e., $\hat{T}_{ij}+\Delta \hat{T}_{ij}\geq0$ for all $i,j$ and $\hat{T}_{ii}+\Delta \hat{T}_{ii}>\hat{T}_{ij}+\Delta \hat{T}_{ij}$ for all $j\neq i$. 
For any $\delta>0$, with probability at least $1-\delta$, we have
\begin{align}
\mathbb{E}[\bar{R}_{n,w}(\hat{T}+\Delta \hat{T},\hat{f})]- \bar{R}_{n,w}(\hat{T}+\Delta \hat{T},\hat{f})\leq \mathbb{E}[\Phi(S)]+CM\sqrt{\frac{\log{1/\delta}}{2n}}.
\end{align}
\end{lemma}
Detailed proof of Lemma \ref{lemma:1} is provided in Section \ref{sec:app1}.

Using the same trick to derive Rademacher complexity \cite{bartlett2002rademacher}, we have
\begin{align}
 \mathbb{E}[\Phi(S)]\leq 2\mathbb{E}\left[\sup_{\Delta T,f}\frac{1}{n}\sum_{i=1}^{n}\sigma_i\frac{{g}_{\bar{Y}_i}(X_i)}{((\hat{T}+\Delta T)^\top{g})_{\bar{Y}_i}(X_i)}\ell(f(X_i),\bar{Y}_i)\right],
\end{align}
where $\sigma_1,\ldots,\sigma_n$ are i.i.d. Rademacher random variables.

We can upper bound the right hand part of the above inequality by the following lemma.
\begin{lemma} \label{lemma:2}
\begin{align}
\mathbb{E}\left[\sup_{\Delta T,f}\frac{1}{n}\sum_{i=1}^{n}\sigma_i\frac{{g}_{\bar{Y}_i}(X_i)}{((\hat{T}+\Delta T)^\top{g})_{\bar{Y}_i}(X_i)}\ell(f(X_i),\bar{Y}_i)\right]\leq \mathbb{E}\left[\sup_{f}\frac{1}{n}\sum_{i=1}^{n}\sigma_i\ell(f(X_i),\bar{Y}_i)\right].
\end{align}
\end{lemma}

Note that Lemma \ref{lemma:2} is not an application of Talagrand Contraction Lemma \cite{ledoux2013probability}. Detailed proof of Lemma \ref{lemma:2} is provided in Section \ref{sec:app2}.

Recall that ${f}=\arg\max_{i\in\{1,\ldots,C\}}{g}_i$ is the classifier, where $g$ is the output of the softmax function, i.e., $g_i(X)=\exp{(h_i(X))}/\sum_{k=1}^{C}\exp{(h_k(X)}, i=1,\ldots,C$, and $h(X)$ is defined by a $d$-layer neural network, i.e., $h: X\mapsto W_d\sigma_{d-1}(W_{d-1}\sigma_{d-2}(\ldots \sigma_1(W_1X)))\in\mathbb{R}^C$,  $W_1,\ldots,W_d$ are the parameter matrices, and $\sigma_1,\ldots,\sigma_{d-1}$ are activation functions. To further upper bound the Rademacher complexity, we need to consider the Lipschitz continuous property of the loss function w.r.t. to $h(X)$. To avoid more assumption, We discuss the widely used \textit{cross-entropy loss}, i.e., 
\begin{align}
 \ell(f(X),\bar{Y})=-\sum_{i=1}^{C}1_{\{\bar{Y}=i\}}\log(g_i(X)).
\end{align}

We can further upper bound the Rademacher complexity by the following lemma.

\begin{lemma} \label{lemma:3}
\begin{align}
\mathbb{E}\left[\sup_{f}\frac{1}{n}\sum_{i=1}^{n}\sigma_i\ell(f(X_i),\bar{Y}_i)\right]\leq CL\mathbb{E}\left[ \sup_{h\in H }\frac{1}{n}\sum_{i=1}^{n}\sigma_ih(X_i)\right],
\end{align}
where $H$ is the function class induced by the deep neural network.
\end{lemma}
Detailed proof of Lemma \ref{lemma:3} is provided in Section \ref{sec:app3}.

Note that $\mathbb{E}\left[ \sup_{h\in H }\frac{1}{n}\sum_{i=1}^{n}\sigma_ih(X_i)\right]$ measures the hypothesis complexity of deep neural networks, which has been widely studied recently \cite{neyshabur2017exploring,bartlett2017spectrally,golowich2018size,neyshabur2018pac}. Specifically, \cite{golowich2018size} proved the following theorem (Theorem 1 therein).
\begin{theorem}\label{thm:network}
Assume the Frobenius norm of the weight matrices $W_1,\ldots,W_d$ are at most $M_1,\ldots, M_d$. Let the activation functions be 1-Lipschitz, positive-homogeneous, and applied element-wise (such as the ReLU). Let $x$ is upper bounded by B, i.e., for any $x
\in \mathcal{X}$, $\|x\|\leq B$. Then,
\begin{align}
\mathbb{E}\left[ \sup_{h\in H }\frac{1}{n}\sum_{i=1}^{n}\sigma_ih(X_i)\right]\leq \frac{B(\sqrt{2d\log2}+1)\Pi_{i=1}^{d}M_i}{\sqrt{n}}.
\end{align}
\end{theorem}

Theorem 1 follows by combining Lemmas 1, 2, 3, and Theorem \ref{thm:network}.

\subsection{Proof of Lemma \ref{lemma:1}}\label{sec:app1}

We employ McDiarmid's concentration inequality \cite{boucheron2013concentration} to prove the lemma. We first check the bounded difference property of $\Phi(S)$, e.g.,
\begin{equation}
\Phi(S)-\Phi(S^i)\leq \sup_{\Delta T,f}\frac{1}{n}\left(\frac{{g}_{\bar{Y}_i}(X_i)\ell(f(X_i),\bar{Y}_i)}{((\hat{T}+\Delta T)^\top{g})_{\bar{Y}_i}(X_i)}-\frac{{g}_{\bar{Y}'_i}(X'_i)\ell(f(X'_i),\bar{Y}'_i)}{((\hat{T}+\Delta T)^\top{g})_{\bar{Y}'_i}(X'_i)}\right).
\end{equation}
Before further upper bounding the above difference, 
we show that the weighted loss is upper bounded by $CM$. Specifically, we have assume the learned transition matrix is valid, i.e., $\hat{T}_{ij}+\Delta {T}_{ij}\geq0$ for all $i,j$ and $\hat{T}_{ii}+\Delta {T}_{ii}>\hat{T}_{ij}+\Delta {T}_{ij}$ for all $j\neq i$. Thus $\frac{{g}_{\bar{Y}}(X)}{((\hat{T}+\Delta T)^\top{g})_{\bar{Y}}(X)} \leq 1/\min_i(\hat{T}_{ii}+\Delta {T}_{ii})\leq C$ for any $(X,\bar{Y})$ and $\hat{g}$. Then, we can conclude that the weighted loss is upper bounded by $CM$ and that
\begin{equation}
\Phi(S)-\Phi(S^i)\leq \frac{CM}{n}.
\end{equation}
Similarly, we could prove that $\Phi(S^i)-\Phi(S)\leq \frac{CM}{n}$.

By employing McDiarmid's concentration inequality, for any $\delta>0$, with probability at least $1-\delta$, we have
\begin{equation}
\Phi(S)-\mathbb{E}[\Phi(S)]\leq CM\sqrt{\frac{\log(1/\delta)}{2n}}.
\end{equation}

\subsection{Proof of Lemma \ref{lemma:2}}\label{sec:app2}
Given the learned transition matrix is valid, we have shown that $\frac{{g}_{\bar{Y}}(X)}{((\hat{T}+\Delta T)^\top{g})_{\bar{Y}}(X)} \leq 1/\min_i(\hat{T}_{ii}+\Delta {T}_{ii})\leq C$ for all $(X,\bar{Y})$ in the proof of Lemma \ref{lemma:1}.

Lemma \ref{lemma:2} holds of we could prove the following inequality
\begin{align}
\mathbb{E}_\sigma\left[\sup_{\Delta T,f}\frac{1}{n}\sum_{i=1}^{n}\sigma_i\frac{{g}_{\bar{Y}_i}(X_i)}{((\hat{T}+\Delta T)^\top{g})_{\bar{Y}_i}(X_i)}\ell(f(X_i),\bar{Y}_i)\right]\leq \mathbb{E}_\sigma\left[\sup_{f}\frac{1}{n}\sum_{i=1}^{n}\sigma_i\ell(f(X_i),\bar{Y}_i)\right].
\end{align}

Note that
\begin{equation}
\begin{aligned}
&\mathbb{E}_\sigma\left[\sup_{\Delta T,f}\frac{1}{n}\sum_{i=1}^{n}\sigma_i\frac{{g}_{\bar{Y}_i}(X_i)}{((\hat{T}+\Delta T)^\top{g})_{\bar{Y}_i}(X_i)}\ell(f(X_i),\bar{Y}_i)\right]\\
&=\mathbb{E}_{\sigma_1,\ldots,\sigma_{n-1}}\left[\mathbb{E}_{\sigma_n}\left[\sup_{\Delta T,f}\frac{1}{n}\sum_{i=1}^{n}\sigma_i\frac{{g}_{\bar{Y}_i}(X_i)}{((\hat{T}+\Delta T)^\top{g})_{\bar{Y}_i}(X_i)}\ell(f(X_i),\bar{Y}_i)\right]\right].
\end{aligned}
\end{equation}

Let $s_{n-1}(\Delta T,f)=\sum_{i=1}^{n-1}\sigma_i\frac{{g}_{\bar{Y}_i}(X_i)}{((\hat{T}+\Delta T)^\top{g})_{\bar{Y}_i}(X_i)}\ell(f(X_i),\bar{Y}_i)$.

By definition of the supremum, for any $\epsilon>0$, there exist $(\Delta T,f_1)$ and $(\Delta T,f_2)$ such that
\begin{equation}
\begin{aligned}
&\frac{{g}_{\bar{Y}_n}(X_n)}{((\hat{T}+\Delta T)^\top{g})_{\bar{Y}_n}(X_n)}\ell(f_1(X_n),\bar{Y}_n)+s_{n-1}(\Delta T,f_1)\\
&\geq (1-\epsilon)\sup_{\Delta T, f}\left(\frac{{g}_{\bar{Y}_n}(X_n)}{((\hat{T}+\Delta T)^\top{g})_{\bar{Y}_n}(X_n)}\ell(f(X_n),\bar{Y}_n)+s_{n-1}(\Delta T,f)\right)
\end{aligned}
\end{equation}
and 
\begin{equation}
\begin{aligned}
&-\frac{{g}_{\bar{Y}_n}(X_n)}{((\hat{T}+\Delta T)^\top{g})_{\bar{Y}_n}(X_n)}\ell(f_2(X_n),\bar{Y}_n)+s_{n-1}(\Delta T,f_2)\\
&\geq (1-\epsilon)\sup_{\Delta T, f}\left(-\frac{{g}_{\bar{Y}_n}(X_n)}{((\hat{T}+\Delta T)^\top{g})_{\bar{Y}_n}(X_n)}\ell(f(X_n),\bar{Y}_n)+s_{n-1}(\Delta T,f)\right).
\end{aligned}
\end{equation}
Thus, for any $\epsilon$, we have
\begin{equation}
\begin{aligned}
&(1-\epsilon)\mathbb{E}_{\sigma_n}\left[\sup_{\Delta T, f}\left(\sigma_n\frac{{g}_{\bar{Y}_n}(X_n)}{((\hat{T}+\Delta T)^\top{g})_{\bar{Y}_n}(X_n)}\ell(f(X_n),\bar{Y}_n)+s_{n-1}(\Delta T,f)\right)\right]\\
&=\frac{(1-\epsilon)}{2}\sup_{\Delta T, f}\left(\frac{{g}_{\bar{Y}_n}(X_n)}{((\hat{T}+\Delta T)^\top{g})_{\bar{Y}_n}(X_n)}\ell(f_1(X_n),\bar{Y}_n)+s_{n-1}(\Delta T,f_1)\right)\\
&+\frac{(1-\epsilon)}{2}\sup_{\Delta T, f}\left(-\frac{{g}_{\bar{Y}_n}(X_n)}{((\hat{T}+\Delta T)^\top{g})_{\bar{Y}_n}(X_n)}\ell(f_2(X_n),\bar{Y}_n)+s_{n-1}(\Delta T,f_2)\right)\\
&\leq\frac{1}{2}\left(\frac{{g}_{\bar{Y}_n}(X_n)}{((\hat{T}+\Delta T)^\top{g})_{\bar{Y}_n}(X_n)}\ell(f_1(X_n),\bar{Y}_n)+s_{n-1}(\Delta T,f_1)\right.\\
&\left.+ (s_{n-1}(\Delta T,f_2)-\frac{{g}_{\bar{Y}_n}(X_n)}{((\hat{T}+\Delta T)^\top{g})_{\bar{Y}_n}(X_n)}\ell(f_2(X_n),\bar{Y}_n)\right)\\
&\leq\frac{1}{2}\left(s_{n-1}(\Delta T,f_1)+ s_{n-1}(\Delta T,f_2)+C|\ell(f_1(X_n),\bar{Y}_n)-\ell(f_2(X_n),\bar{Y}_n)|\right),
\end{aligned}
\end{equation}
where the last inequality holds because $\frac{{g}_{\bar{Y}}(X)}{((\hat{T}+\Delta T)^\top{g})_{\bar{Y}}(X)}\leq C$ for any $(X,\bar{Y}),{g}$, and valid $\hat{T}+\Delta T$.

Let $s=\text{sgn}(\ell(f_1(X_n),\bar{Y}_n)-\ell(f_2(X_n),\bar{Y}_n))$. We have
\begin{equation}
\begin{aligned}
&(1-\epsilon)\mathbb{E}_{\sigma_n}\left[\sup_{\Delta T, f}\left(\sigma_n\frac{{g}_{\bar{Y}_n}(X_n)}{((\hat{T}+\Delta T)^\top{g})_{\bar{Y}_n}(X_n)}\ell(f(X_n),\bar{Y}_n)+s_{n-1}(\Delta T,f)\right)\right]\\
&\leq\frac{1}{2}\left(s_{n-1}(\Delta T,f_1)+ s_{n-1}(\Delta T,f_2)+sC(\ell(f_1(X_n),\bar{Y}_n)-\ell(f_2(X_n),\bar{Y}_n))\right)\\
&=\frac{1}{2}\left(s_{n-1}(\Delta T,f_1)+sC\ell(f_1(X_n),\bar{Y}_n) \right)+\frac{1}{2}\left(s_{n-1}(\Delta T,f_2)-sC\ell(f_2(X_n),\bar{Y}_n)\right)\\
&\leq \frac{1}{2}\sup_{f\in F}\left(s_{n-1}(\Delta T,f)+sC\ell(f(X_n),\bar{Y}_n) \right)+\frac{1}{2}\sup_{f\in F}\left(s_{n-1}(\Delta T,f)-sC\ell(f(X_n),\bar{Y}_n)\right)\\
&=\mathbb{E}_{\sigma_n}\left[\sup_{\Delta T, f}\left(\sigma_n\ell(f(X_n),\bar{Y}_n)+s_{n-1}(\Delta T,f)\right)\right].
\end{aligned}
\end{equation}
Since the above inequality holds for any $\epsilon>0$, we have
\begin{equation}
\begin{aligned}
&\mathbb{E}_{\sigma_n}\left[\sup_{\Delta T, f}\left(\sigma_n\frac{{g}_{\bar{Y}_n}(X_n)}{((\hat{T}+\Delta T)^\top{g})_{\bar{Y}_n}(X_n)}\ell(f(X_n),\bar{Y}_n)+s_{n-1}(\Delta T,f)\right)\right]\\
&\leq \mathbb{E}_{\sigma_n}\left[\sup_{\Delta T, f}\left(\sigma_n\ell(f(X_n),\bar{Y}_n)+s_{n-1}(\Delta T,f)\right)\right].
\end{aligned}
\end{equation}

Proceeding in the same way for all other $\sigma$, we have
\begin{equation}
\begin{aligned}
&\mathbb{E}_{\sigma}\left[\sup_{\Delta T, f}\sum_{i=1}^n\sigma_i\frac{{g}_{\bar{Y}_i}(X_i)}{((\hat{T}+\Delta T)^\top{g})_{\bar{Y}_i}(X_i)}\ell(f(X_i),\bar{Y}_i)\right]\leq \mathbb{E}_{\sigma}\left[\sup_{f\in F}\sum_{i=1}^n\sigma_i\ell(f(X_i),\bar{Y}_i)\right].
\end{aligned}
\end{equation}
and thus
\begin{equation}
\begin{aligned}
&\mathbb{E}\left[\sup_{\Delta T, f}\sum_{i=1}^n\sigma_i\frac{{g}_{\bar{Y}_i}(X_i)}{((\hat{T}+\Delta T)^\top{g})_{\bar{Y}_i}(X_i)}\ell(f(X_i),\bar{Y}_i)\right]\leq \mathbb{E}\left[\sup_{f\in F}\sum_{i=1}^n\sigma_i\ell(f(X_i),\bar{Y}_i)\right].
\end{aligned}
\end{equation}

\subsection{Proof of Lemma \ref{lemma:3}}\label{sec:app3}
Before proving Lemma \ref{lemma:3}, we show that the loss function $\ell(f(X),\bar{Y})$ is 1-Lipschitz-continuous w.r.t. $h_i(X),i=\{1,\ldots,C\}$.

Recall that 
\begin{equation}
{\ell}(f(X),\bar{Y}) = -\sum_{i=1}^{C}1_{\{\bar{Y}=i\}}\log(g_i(X))= -\log\left(\frac{\exp(h_{\bar{Y}}(X))}{\sum_{i=1}^C \exp(h_i(X))}\right).
\end{equation}
Take the derivative of $\ell(f(X),\bar{Y})$ w.r.t. $h_i(X)$. If $i\neq \bar{Y}$, we have 
\begin{equation} \label{derivative1}
\begin{aligned}
&\frac{\partial {\ell}(f(X),\bar{Y})}{\partial h_i(X)} = \frac{\exp(h_i(X))}{\sum_{i=1}^c \exp(h_i(X))}.
\end{aligned}
\end{equation}
If $i= \bar{Y}$, we have 
\begin{equation} \label{derivative2}
\begin{aligned}
&\frac{\partial {\ell}(f(X),\bar{Y})}{\partial h_i(X)} = -1+ \frac{\exp(h_i(X))}{\sum_{i=1}^c \exp(h_i(X))}.
\end{aligned}
\end{equation}

According to Eqs.(\ref{derivative1}) and (\ref{derivative2}), it is easy to conclude that $-1 \leq \frac{\partial {\ell}(f(X),\bar{Y})}{\partial h_i(X)} \leq 1$, which also indicates that the loss function is 1-Lipschitz with respect to $h_i(X), \forall i \in \{1,\ldots,C\}$.

Now we are ready to prove Lemma 3. We have
\begin{equation}
\begin{aligned}
&\mathbb{E}\left[\sup_{f}\frac{1}{n}\sum_{i=1}^{n}\sigma_i\ell(f(X_i),\bar{Y}_i)\right]\\
&= \mathbb{E}\left[\sup_{f=\arg\max\{h_1,\ldots,h_c\} }\frac{1}{n}\sum_{i=1}^{n}\sigma_i{\ell}(f(X_i),\bar{Y}_i)\right] \\
&= \mathbb{E}\left[\sup_{\max\{h_1,\ldots,h_c\} }\frac{1}{n}\sum_{i=1}^{n}\sigma_i{\ell}(f(X_i),\bar{Y}_i)\right] \\
&\leq \mathbb{E}\left[ \sum_{k=1}^C \sup_{h_k\in H }\frac{1}{n}\sum_{i=1}^{n}\sigma_i{\ell}(f(X_i),\bar{Y}_i)\right] \\
&= \sum_{k=1}^C \mathbb{E}\left[ \sup_{h_k\in H }\frac{1}{n}\sum_{i=1}^{n}\sigma_i{\ell}(f(X_i),\bar{Y}_i)\right] \\
&\leq CL\mathbb{E}\left[ \sup_{h_k\in H }\frac{1}{n}\sum_{i=1}^{n}\sigma_ih_k(X_i)\right] \\
&=CL\mathbb{E}\left[ \sup_{h\in H }\frac{1}{n}\sum_{i=1}^{n}\sigma_ih(X_i)\right],
\end{aligned}
\end{equation}
where the first equation holds because the softmax function preserves the rank of its inputs, i.e., $f(X)=\arg\max_{i \in \{1,\ldots,C\}} g_i(X) = \arg\max_{i \in \{1,\ldots,C\}} h_i(X)$; the second equation holds because $\arg\max \{h_1,\cdots,h_c\}$ and $\max\{h_1,\cdots,h_c\}$ give the same constraint on $h_i, \forall i \in\{1,\ldots,C\}$;the fifth inequality holds because of the Talagrand Contraction Lemma \cite{ledoux2013probability}.

\section{Definition of transition matrix}
\label{app:noise}

The definition of symmetry flipping transition matrix is as follows, where $C$ is number of the class.

\begin{align*}
\text{sym-$\epsilon$:}
& \quad
T =
\begin{bmatrix}
1-\epsilon & \frac{\epsilon}{C-1} & \dots  & \frac{\epsilon}{C-1} & \frac{\epsilon}{C-1}\\
\frac{\epsilon}{C-1} & 1-\epsilon & \frac{\epsilon}{C-1} & \dots & \frac{\epsilon}{C-1}\\
\vdots &  & \ddots &  & \vdots\\
\frac{\epsilon}{C-1} & \dots & \frac{\epsilon}{C-1} & 1-\epsilon & \frac{\epsilon}{C-1}\\
\frac{\epsilon}{C-1} & \frac{\epsilon}{C-1} & \dots  & \frac{\epsilon}{C-1} & 1-\epsilon
\end{bmatrix}.
\end{align*}

\section{More discussions about Figure 3}
We represent Figure 3 in the main paper as Figure 1 in this appendix. 

From the figure, we can compare the transition matrices learned by the proposed T-revision method and the traditional anchor point based method. Specifically, as shown in Figure 1, at epoch 0, the estimation error corresponds to the estimation error of transition matrix learned by identifying anchor points \cite{thekumparampil2018robustness} (the traditional method to learn transition matrix). Note that the method with "-N/A" in its name means it runs on the modified datasets where instances with large clean class posterior probobilities are removed (anchor points are removed); while the method with "-A" in its name means it runs the original intact dataset (may contain anchor points). Clearly, we can see that the estimation error will increase by removing possible anchor points, meaning that anchor points is crucial in the traditional transition matrix learning. Moreover, as the number of epochs grows, the figures show how the estimation error varies by running the proposed revision methods. We can see that the proposed Reweight method always leads to smaller estimation errors, showing that the proposed method works well in find a better transition matrix.

Figure 1 also shows the comparison of learning transition matrices between the risk-consistent estimator based method and the classifier-consistent method based method. For classifier-consistent algorithms, we can also modify the transition matrix by adding a slack variable and learning it jointly with the classifier, e.g., Forward-A-R and Forward-N/A-R. However, we can find that the classifier-consistent algorithm based method Forward-N/A-R may fail in learning a good transition matrix, e.g., Figure 1(a). This is because there is no reason to learn the transition matrix by minimizing the classifier-consistent objective function. It is reasonable to learn the transition matrix by minimizing the risk-consistent estimator because a favorable transition matrix should make the classification risk w.r.t. clean data small. This is verified by comparing Forward-A-R and Forward-N/A-R with the proposed Reweight-A-R and Reweight-N/A-R, we can find that the risk-consistent estimator Reweight always leads to smaller estimation errors for learning transition matrix.

\begin{figure}[t]
\centering
\vspace{-5px}
\includegraphics[width=0.32\textwidth]{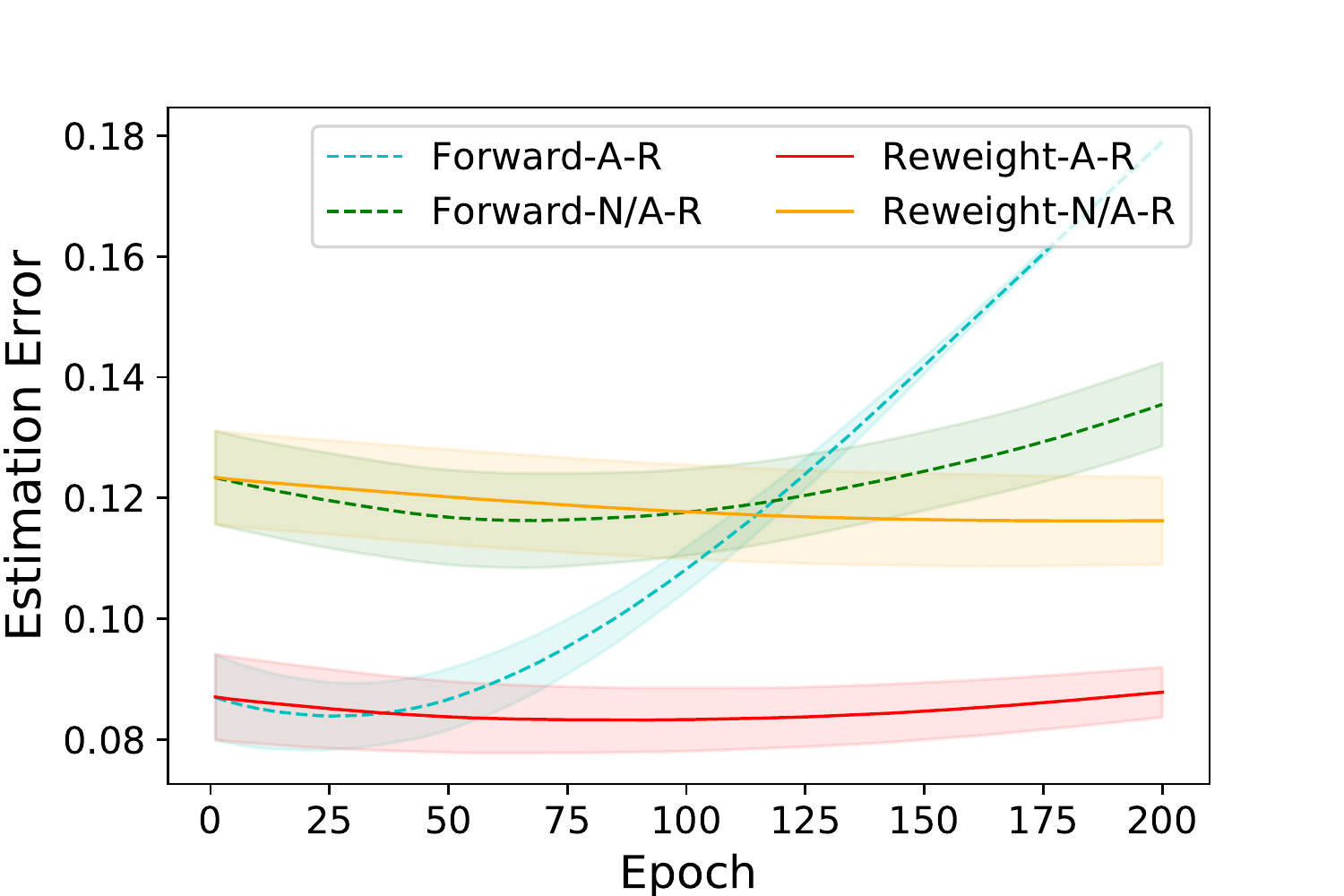}
\includegraphics[width=0.32\textwidth]{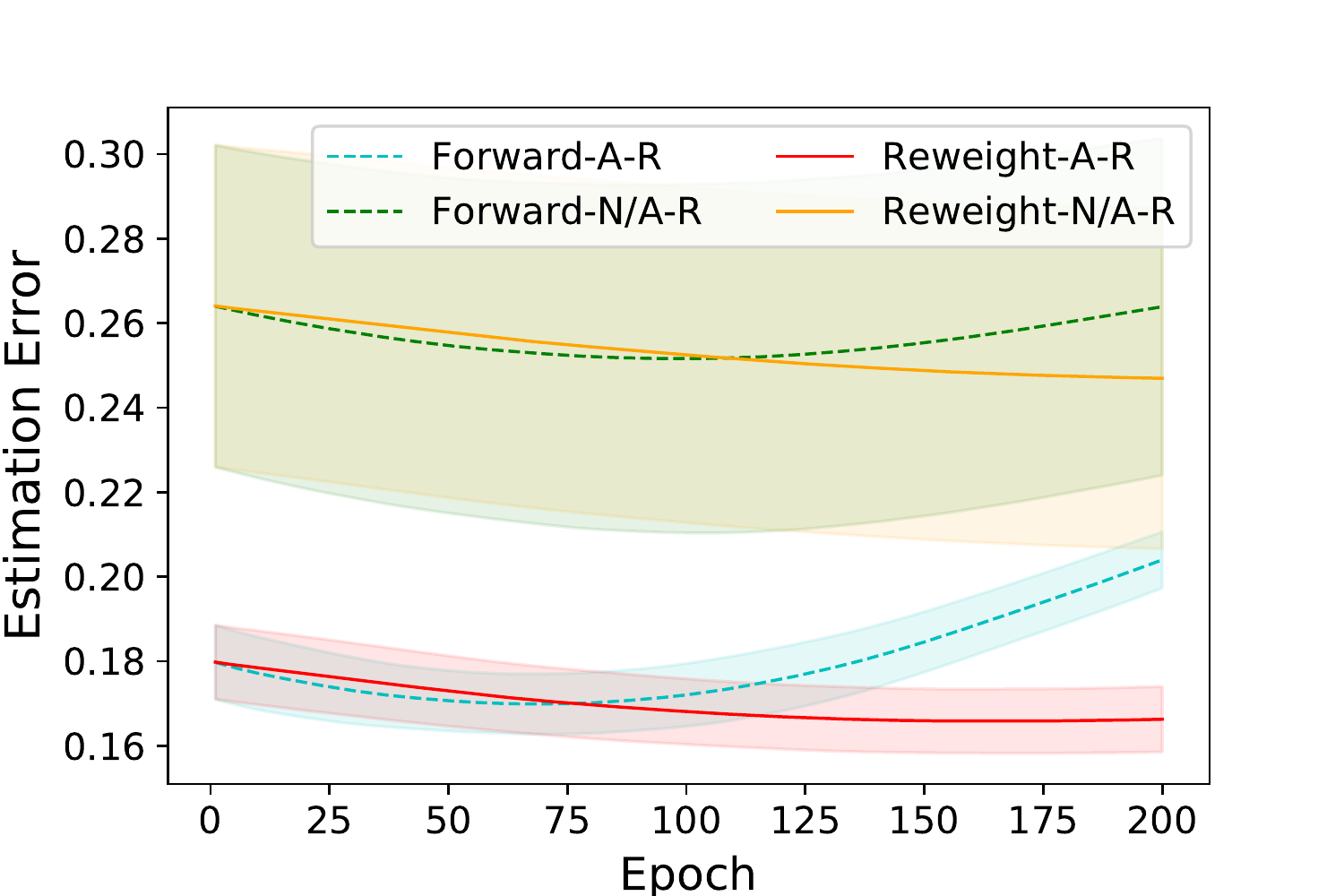}
\includegraphics[width=0.32\textwidth]{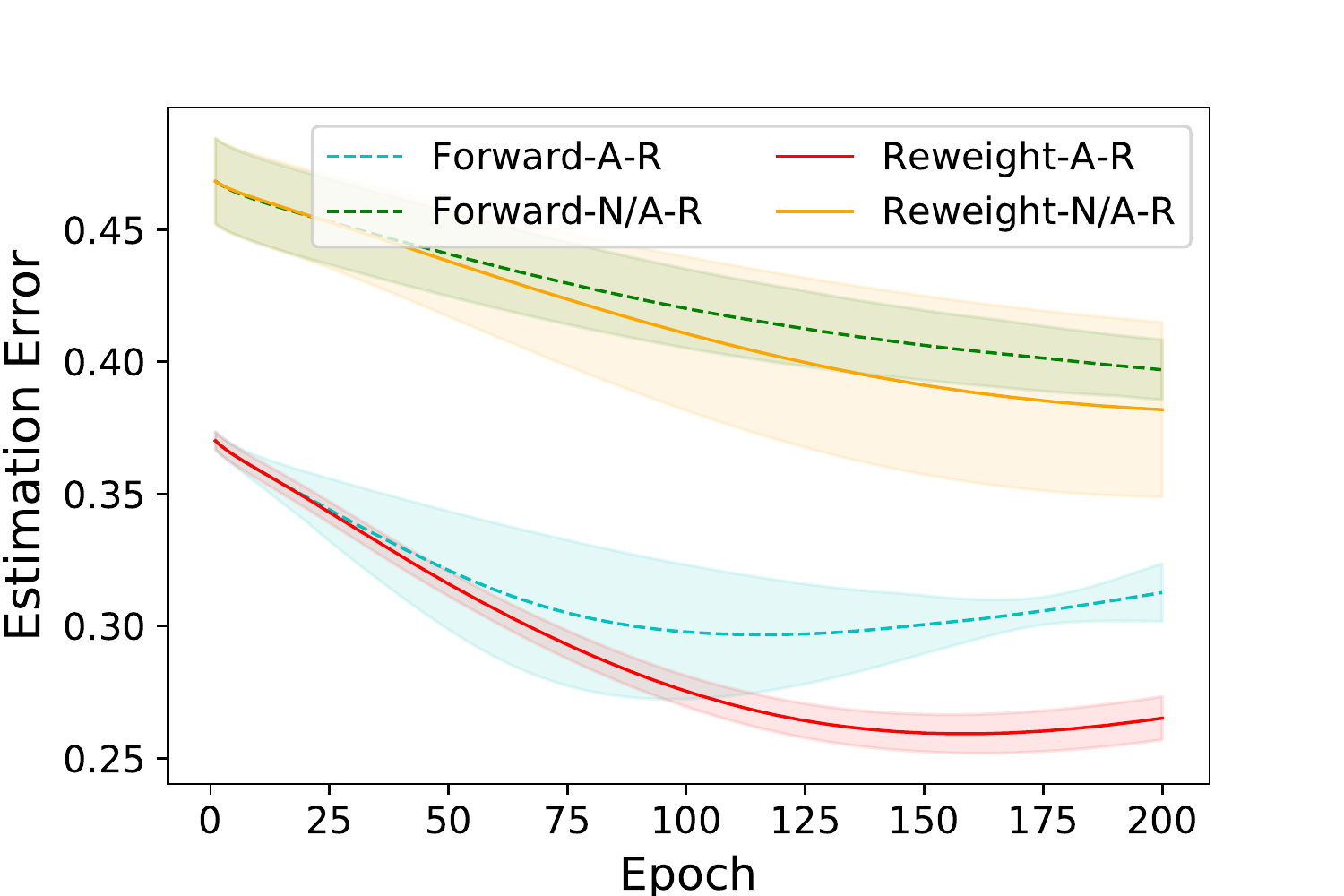}
\subfigure[\textit{MNIST}]
{\includegraphics[width=0.32\textwidth]{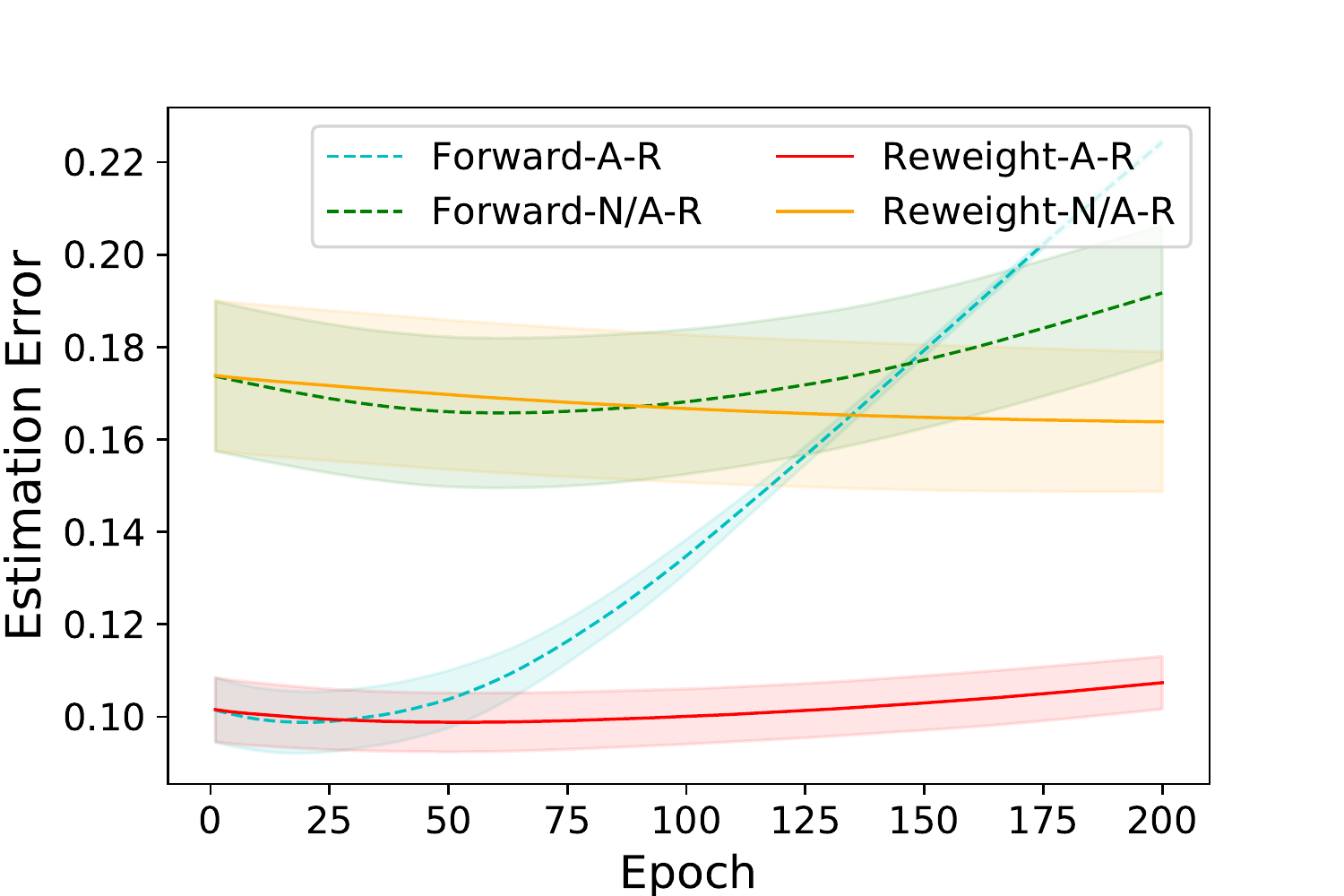}}
\subfigure[\textit{CIFAR-10}]
{\includegraphics[width=0.32\textwidth]{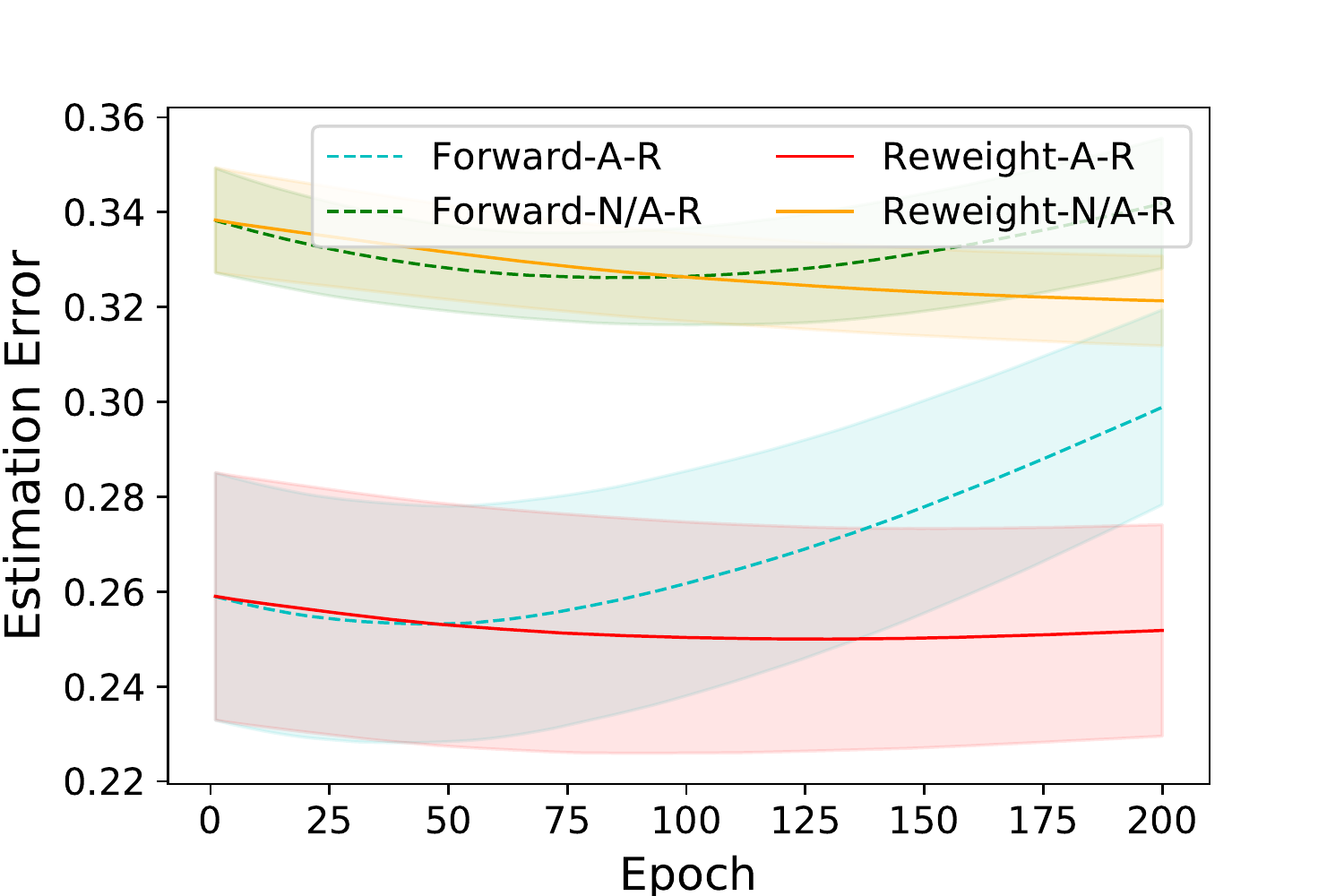}}
\subfigure[\textit{CIFAR-100}]
{\includegraphics[width=0.32\textwidth]{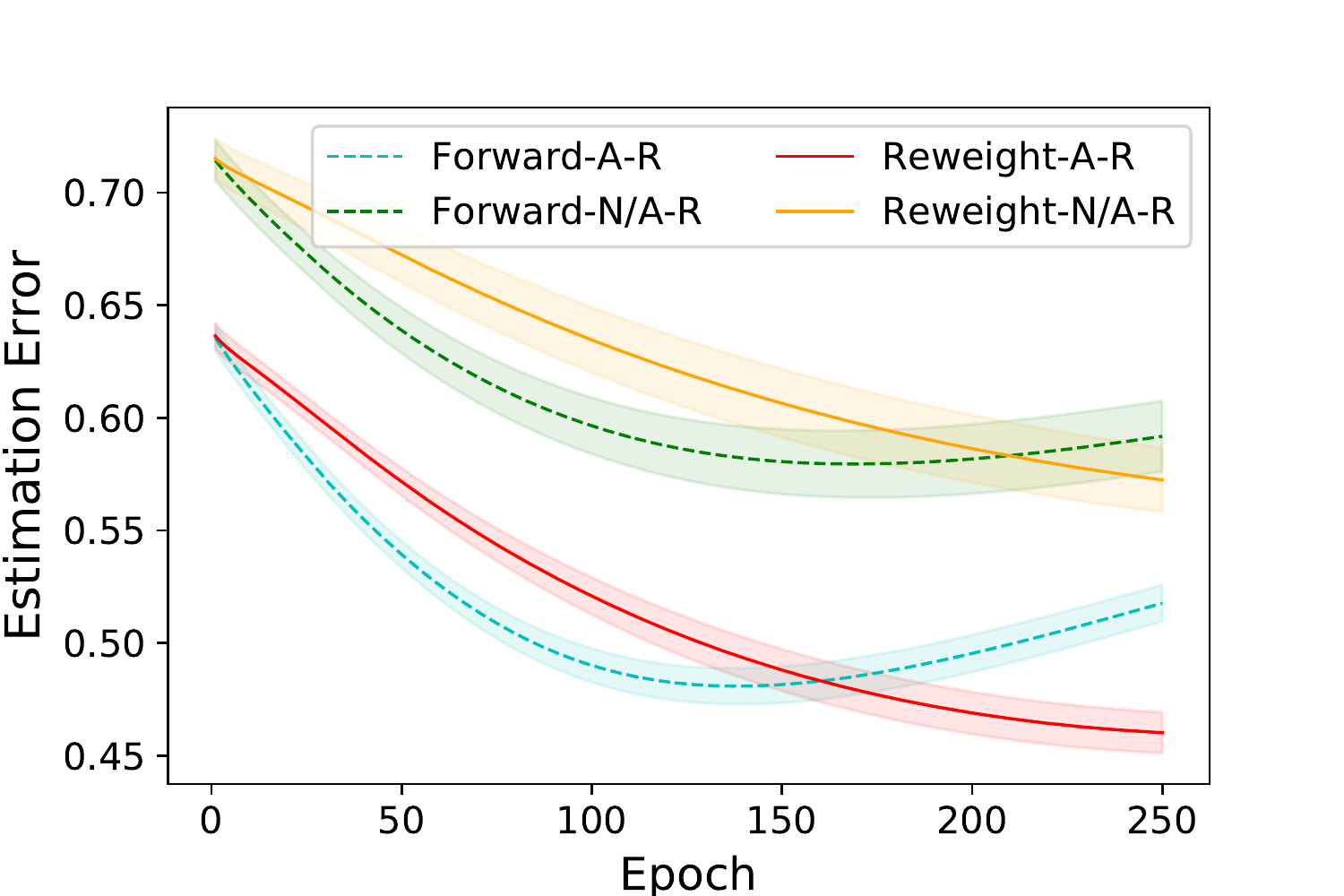}}
\vspace{-10px}
\caption{Comparing the estimation error of the transition matrix by employing classifier-consistent and risk-consistent estimators. The first row is about sym-20 label noise while the second row is about sym-50 label noise. The error bar for STD in each figure has been highlighted as a shade. }
\label{fig:estimation}
\vspace{-5px}
\end{figure}

\end{document}